\documentclass[runningheads]{llncs}

\usepackage{eccv}

\usepackage{eccvabbrv}

\usepackage{graphicx}
\usepackage{booktabs}

\usepackage{multirow}
\usepackage{makecell}
\usepackage{marvosym, ifsym} 

\usepackage[accsupp]{axessibility}  

\usepackage[colorlinks]{hyperref}

\usepackage{orcidlink}

\begin{document}

\title{FreeInit: Bridging Initialization Gap in Video Diffusion Models} 


\author{Tianxing Wu\orcidlink{0000-0001-7345-0254} \and
Chenyang Si\orcidlink{0000-0002-3354-1968} \and
Yuming Jiang\orcidlink{0000-0001-7653-4015} \and
Ziqi Huang\orcidlink{0000-0001-8008-5873} \and
Ziwei Liu\textsuperscript{\Letter}\orcidlink{0000-0002-4220-5958}}

\authorrunning{T.~Wu et al.}


\institute{S-Lab, Nanyang Technological University \\
\email{\{tianxing001, chenyang.si, yuming002, ziqi002, ziwei.liu\}@ntu.edu.sg}
}

\maketitle

\begin{center}
    \url{https://tianxingwu.github.io/pages/FreeInit/}
\end{center}

\begin{abstract}
  Though diffusion-based video generation has witnessed rapid progress, the inference results of existing models still exhibit unsatisfactory temporal consistency and unnatural dynamics. In this paper, we delve deep into the noise initialization of video diffusion models, and discover an implicit training-inference gap that attributes to the unsatisfactory inference quality.
  Our key findings are: \textbf{1)} the spatial-temporal frequency distribution of the initial noise at inference is intrinsically different from that for training, and \textbf{2)} the denoising process is significantly influenced by the low-frequency components of the initial noise. Motivated by these observations, we propose a concise yet effective inference sampling strategy, \textbf{FreeInit}, which significantly improves temporal consistency of videos generated by diffusion models. Through iteratively refining the spatial-temporal low-frequency components of the initial latent during inference, FreeInit is able to compensate the initialization gap between training and inference, thus effectively improving the subject appearance and temporal consistency of generation results. Extensive experiments demonstrate that FreeInit consistently enhances the generation quality of various text-to-video diffusion models without additional training or fine-tuning.
\end{abstract}

\section{Introduction}
\label{sec:intro}

\begin{figure}[tb]
  \centering
  \includegraphics[width=1.0\textwidth]{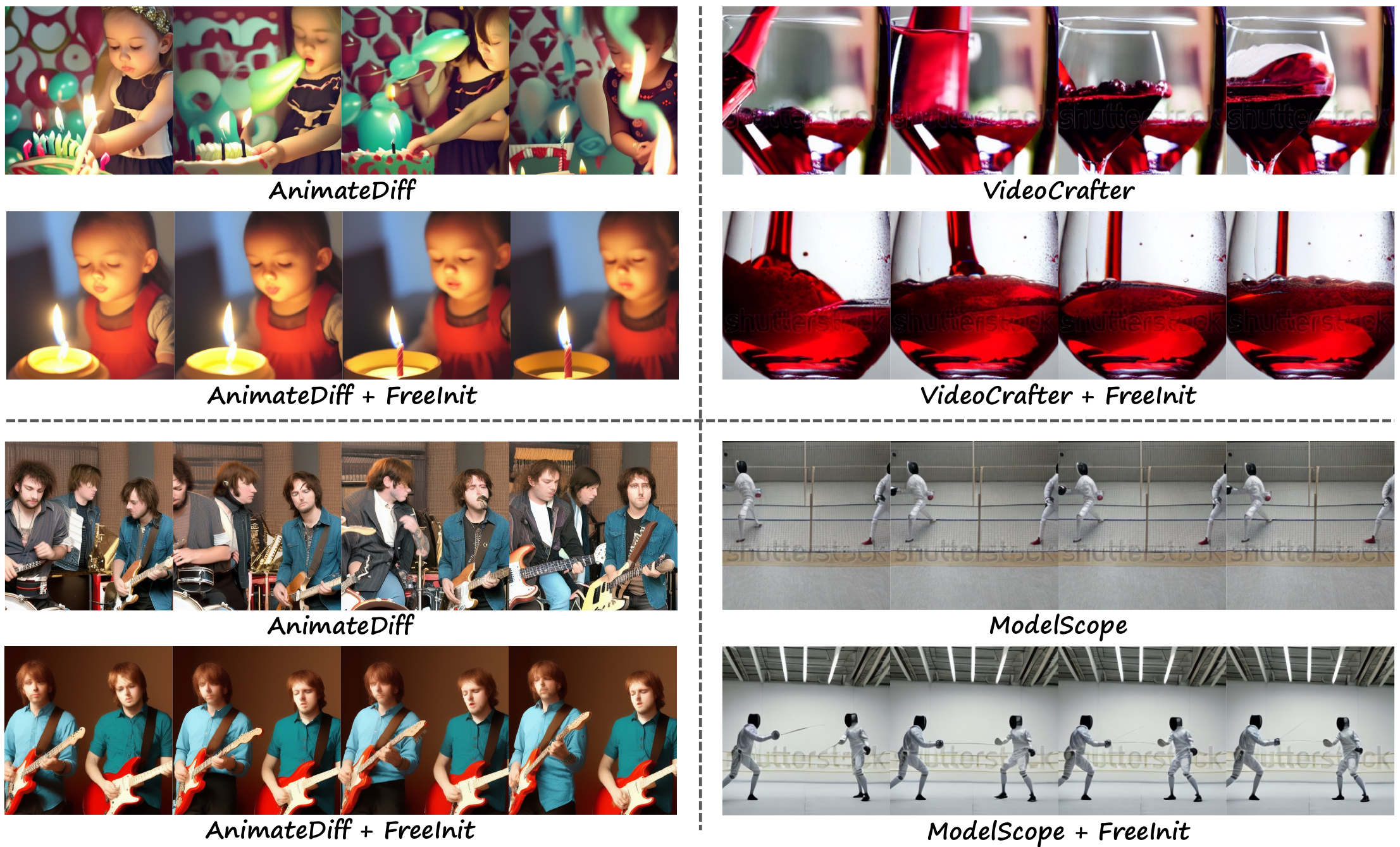}
  \caption{
  \textbf{FreeInit for Video Generation}. We propose \textit{\textbf{FreeInit}}, a concise yet effective method to significantly improve temporal consistency of videos generated by diffusion models. FreeInit requires no additional training, introduces no learnable parameters, and can be easily incorporated into arbitrary video diffusion models at inference time.
  }
  \label{fig:teaser}
\end{figure}

Recently, diffusion models have demonstrated impressive generative capabilities in text-to-image generation~\cite{rombach2022ldm, ramesh2022dalle2, saharia2022imagen}. These advancements have attracted substantial attention, highlighting the potential of creating diverse and realistic images based on textual descriptions. In light of these achievements, researchers are now exploring the application of diffusion models in text-to-video (T2V) generation~\cite{singer2022makeavideo, ho2022imagenvideo, he2022lvdm, zhou2022magicvideo, wang2023modelscope, blattmann2023videoldm, guo2023animatediff, wang2023lavie, chen2023videocrafter1, zhang2023show1}, with the goal of synthesizing visually appealing and contextually coherent videos from textual descriptions. Most of these video diffusion models are built upon powerful pretrained image diffusion models, \eg, Stable Diffusion (SD)~\cite{rombach2022ldm}. Through the incorporation of temporal layers and large-scale training on extensive video datasets, these models are capable of generating video clips that align with the given text prompts. Despite these advancements, the videos generated by these models often suffer from issues related to temporal inconsistency and unnatural dynamics.

\begin{figure}[t]
   \centering
  \includegraphics[width=1.0\linewidth]{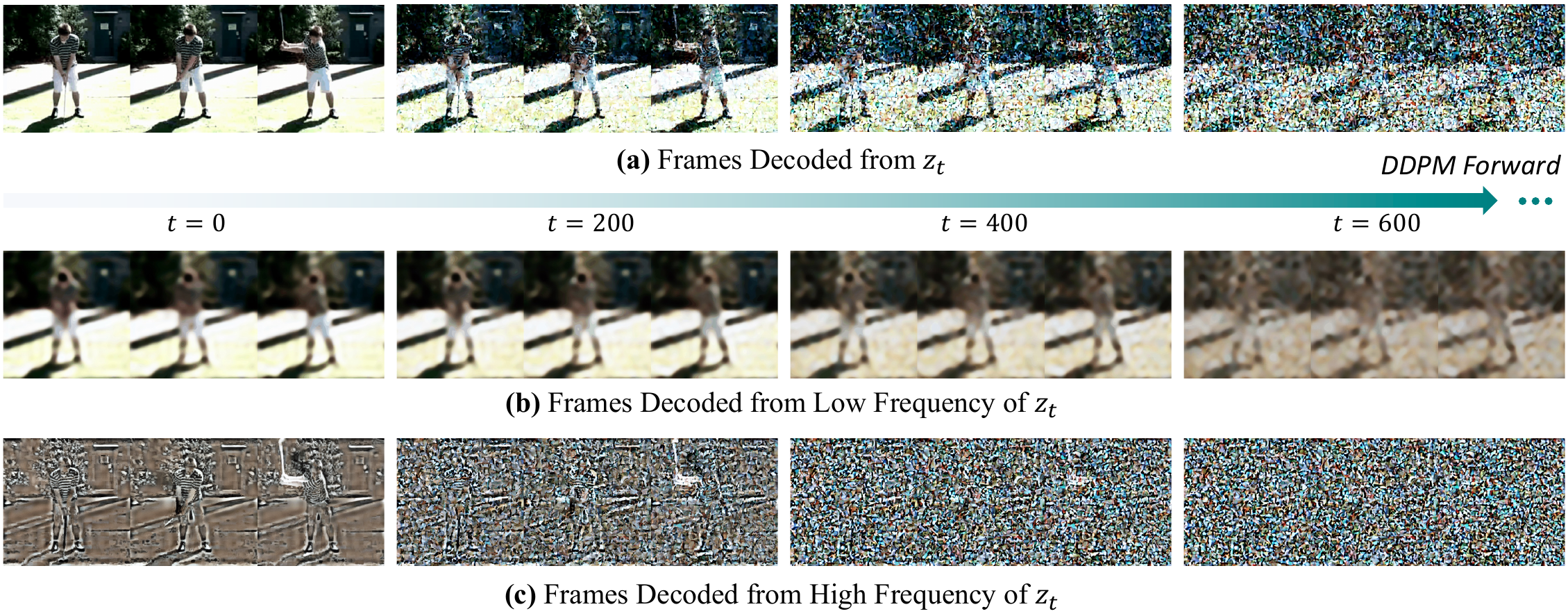}
  \caption{
   \textbf{ Visualization of Decoded Noisy Latent from Different Spatio-Temporal Frequency Bands at Training.} 
   (a) Video frames decoded from the entire frequency band of the noisy latent $z_t$ in DDPM Forward Process. 
   (b) Frames decoded from the low-frequency components of $z_t$. It is evident that the diffusion process has difficulty in fully corrupting the semantics, leaving substantial spatio-temporal correlations in the low-frequency components.
   (c) Frames decoded from the high-frequency components
   of $z_t$. Each frame degenerates rapidly with the diffusion process.
   }
   \label{fig:decomposition}
\end{figure}

In this paper, we delve into the impact of noise initialization on video generation, identifying a significant disparity between the training and the inference process. 
Specifically, we find that the diffusion process fails to fully corrupt the clean latent into pure Gaussian noise, especially in the low-frequency band. To illustrate this, Figure \ref{fig:decomposition} shows the frames decoded from noisy latent during the diffusion process, alongside a spatio-temporal frequency decomposition to assess the extent of corruption across different frequency bands. Remarkably, the corruption of low-frequency components occurs at a notably slower rate than that of the high-frequency components. As a result, the noisy latent at the final diffusion step ($t$=1000) will still contain considerable low-frequency information from the input video. Since the real video frames are temporally correlated in nature, this information leakage eventually leads to an implicit gap between training and inference: at training, the initial noises corrupted from real videos remain temporally correlated at low-frequency band, while during inference, the i.i.d Gaussian initial noise is entirely uncorrelated.
Furthermore, we discover that these low-frequency components can substantially impact the quality of the generated videos, as revealed in our observations in \cref{fig:influence,fig:naive}. 
Thus, when applying the diffusion models trained with the correlated initial noises to non-correlated Gaussian initial noise at inference, the performance deteriorates, exhibiting unsatisfactory temporal consistency and unnatural motions.

Motivated by these observations, we propose a novel inference-time sampling method, denoted as \textbf{\textit{FreeInit}}, to bridge the initialization gap between training and inference without any additional training or fine-tuning. Specifically, during the inference process, 
we first initialize an independent Gaussian noise, which then undergoes the DDIM denoising process to generate a clean video latent.
Subsequently, we obtain a noisy version
of the clean video latent through the forward diffusion process.
Since this noisy latent is obtained from the denoised latent rather than pure noise, its low-frequency components have improved temporal consistency.
With this noisy latent, 
we proceed to reinitialize the noise by combining its low-frequency components with the high-frequency components from a random Gaussian noise using spatio-temporal frequency filter. Finally, this reinitialized noise serves as the starting point for a new round of DDIM sampling.
By iterating this refinement process several times, the initial noise at inference is gradually guided toward the training distribution, facilitating the generation of frames with enhanced temporal consistency and visual appearance.

Extensive experiments across diverse evaluation prompt sets demonstrate the steady enhancement 
brought about by FreeInit for various text-to-video generation models. As illustrated in \cref{fig:teaser}, FreeInit plays a significant role in improving temporal consistency and the visual appearance of generated frames. This method can be readily applied during inference without the need for parameter tuning. Furthermore, to achieve superior generation quality, the frequency filter can be conveniently adjusted for each customized base model.
We summarize our contributions as follows:
\begin{itemize}
    \item We systematically investigate the noise initialization of video diffusion models, and identify an implicit training-inference gap that contributes to the inference quality drop. To our knowledge, we are the first to study the impact of initial noise on video diffusion models from the frequency domain.
    \item We propose a concise yet effective sampling strategy, referred to as FreeInit, which iteratively refines the initial noise without the need for additional training or fine-tuning.
    \item Extensive quantitative and qualitative experiments demonstrate that FreeInit can be effectively applied to various text-to-video models. It consistently improves the inference quality of generated videos.
\end{itemize}

\section{Related Work}
\label{sec:related}

\noindent \textbf{Video Generative Models.} 
There are mainly three types of video generation models, namely GAN-based~\cite{goodfellow2014gan}, transformer-based~\cite{vaswani2017attention}, and diffusion-based~\cite{ho2020ddpm}.
StyleGAN-V~\cite{skorokhodov2022styleganv}, MoCoGAN-HD~\cite{tian2021good}, and \cite{brooks2022generating} utilize the powerful StyleGAN~\cite{karras2019stylegan1, karras2020stylegan2, karras2021stylegan3} to generate videos.
Transformer-based models~\cite{hong2022cogvideo, villegas2022phenaki, wu2021nuwa, wu2021godiva, jiang2023text2performer} such as Phenaki~\cite{villegas2022phenaki}, CogVideo~\cite{hong2022cogvideo}, and NÜWA~\cite{wu2021nuwa} encode videos as visual tokens and train transformer models to auto-regressively generate the visual tokens. 
Recently, diffusion models~\cite{sohl2015deep, song2020score, ho2020ddpm, dhariwal2021beatgan} have made remarkable progress in text-to-image generation~\cite{nichol2021glide, saharia2022imagen,  rombach2022ldm, podell2023sdxl}, and have enabled a line of works that extends these pre-trained diffusion models towards text-to-video generation~\cite{ zhou2022magicvideo, khachatryan2023text2videozero, zhang2023show1, ge2023pyoco, wang2023modelscope, luo2023videofusion,   singer2022makeavideo, ho2022imagenvideo, blattmann2023videoldm, guo2023animatediff, harvey2022fdm, ho2022videoDM, wang2023lavie, he2022lvdm, zhou2023magicvideo}. 
In this work, our method is built on top of diffusion-based text-to-video methods. We propose to iteratively refine the initial noise to improve temporal consistency of pre-trained video diffusion models.
We demonstrate the effectiveness of our method on various diffusion models, including VideoCrafter, ModelScopeT2V (denoted as ModelScope), and AnimateDiff.
VideoCrafter~\cite{he2022lvdm} employs the pre-trained text-to-image model Stable Diffusion~\cite{rombach2022ldm} and incorporates newly initialized temporal layers to enable video generation.
ModelScope~\cite{wang2023modelscope} also initializes the
spatial part from Stable Diffusion and adds spatio-temporal block to learn temporal dependencies.
AnimateDiff~\cite{guo2023animatediff} trains motion modeling modules and inserts them into personalized text-to-image diffusion models to achieve animated videos of customized concepts, \eg, characters, styles, \etc.

\noindent \textbf{Noise in Diffusion Models.} 
Only a few previous works have mentioned the limitations of the noise schedule of current diffusion models. 
In the image domain,~\cite{lin2023common} points out common diffusion noise schedules cannot fully corrupt information in natural images, limiting the model to only generate images with medium brightness. A rescaled training schedule is then proposed to alleviate this problem through fine-tuning.
Recently,~\cite{everaert2023exploiting} makes further discussions on the signal leakage issue, and proposes to explicitly model the signal leakage for better inference noise distribution, which produces images with more diverse brightness and colours.
A resampling operation similar to our iterative refinement strategy is proposed in~\cite{lugmayr2022repaint} to harmonize the inpainted image across full inference timesteps. Different from this, we tackle the initialization problem and explore in frequency-domain to improve temporal consistency.
In the video domain, PYoCo~\cite{ge2023pyoco} carefully designs the progressive video noise prior to achieve a better video generation performance. 
Similar to~\cite{lin2023common}, PYoCo also focuses on the noise schedule at training stage and requires massive fine-tuning on video datasets. In contrast, we focus on the initial noise at inference stage and propose a concise inference-time sampling strategy that bridges the training-inference discrepancy with no fine-tuning required. Some recent works~\cite{gu2023reuse, qiu2023freenoise} also pay attention to the inference initial noise, but aiming at generating long videos. We instead focus on improving inference quality, and further design specific frequency-domain-based operations to modulate different frequency components of the initial noise.

\section{Preliminaries and Observations}

\subsection{Preliminaries}

Similar to image diffusion models, \textit{training} video diffusion models also involve a diffusion process and a denoising process, and operate in the latent space of an autoencoder. The diffusion process includes a sequence of $T$ steps. At each step $t$, Gaussian noise is incrementally added to the video latent $\boldsymbol{z}_0 $, following a predefined variance schedule $\beta_1, \ldots, \beta_T$:
\begin{align}
    q(z_{1:T}|z_0) &= \prod_{t=1}^{T}q(z_{t}|z_{t-1}), \\
    q(z_{t}|z_{t-1}) &= \mathcal{N} (z_{t}; \sqrt{1-\beta_{t}}z_{t-1},\beta_{t}\mathbf{I}).
\end{align}
Let $\alpha_t={1-\beta_t}$, $\overline{\alpha_t}=\prod_{s=1}^{t}\alpha_s$:
\begin{gather}
    q(z_{t}|z_{0}) = \mathcal{N} (z_{t}; \sqrt{\overline{\alpha}_t}z_{0},(1-\overline{\alpha}_t)\mathbf{I}).
\end{gather}
As a result, the noisy latent $z_t$ at each timestep $t$ can be directly sampled as:
\begin{gather}
    z_t = \sqrt{\overline{\alpha}_t} z_0 + \sqrt{1-\overline{\alpha}_t}\epsilon ,
    \label{eq:training-init}
\end{gather}
where $\epsilon \sim \mathcal{N}(\mathbf{0},\mathbf{I})$ is a Gaussian white noise with the same shape as $z_t$.

In the reverse process, the network learns to recover the clean latent $z_0$ by iterative denoising with U-Net~\cite{ronneberger2015unet}, starting from the initial noise $z_T$:
\begin{align}
    p_{\theta}(z_{0:T}) &= p(z_T)\prod_{t=1}^{T} p_{\theta}(z_{t-1} | z_{t}), \\
    p_{\theta}(z_{t-1} | z_{t}) &= \mathcal{N} (z_{t-1}; \mu_\theta(z_t, t), \Sigma_\theta(z_t, t)),
\end{align}
where $\mu_\theta$ and $\Sigma_\theta$ are predicted by the denoising U-Net $\epsilon_\theta$.

\begin{figure}[t]
    \centering
    \begin{minipage}[t]{.48\textwidth}
        \centering
        \includegraphics[width=1\textwidth]{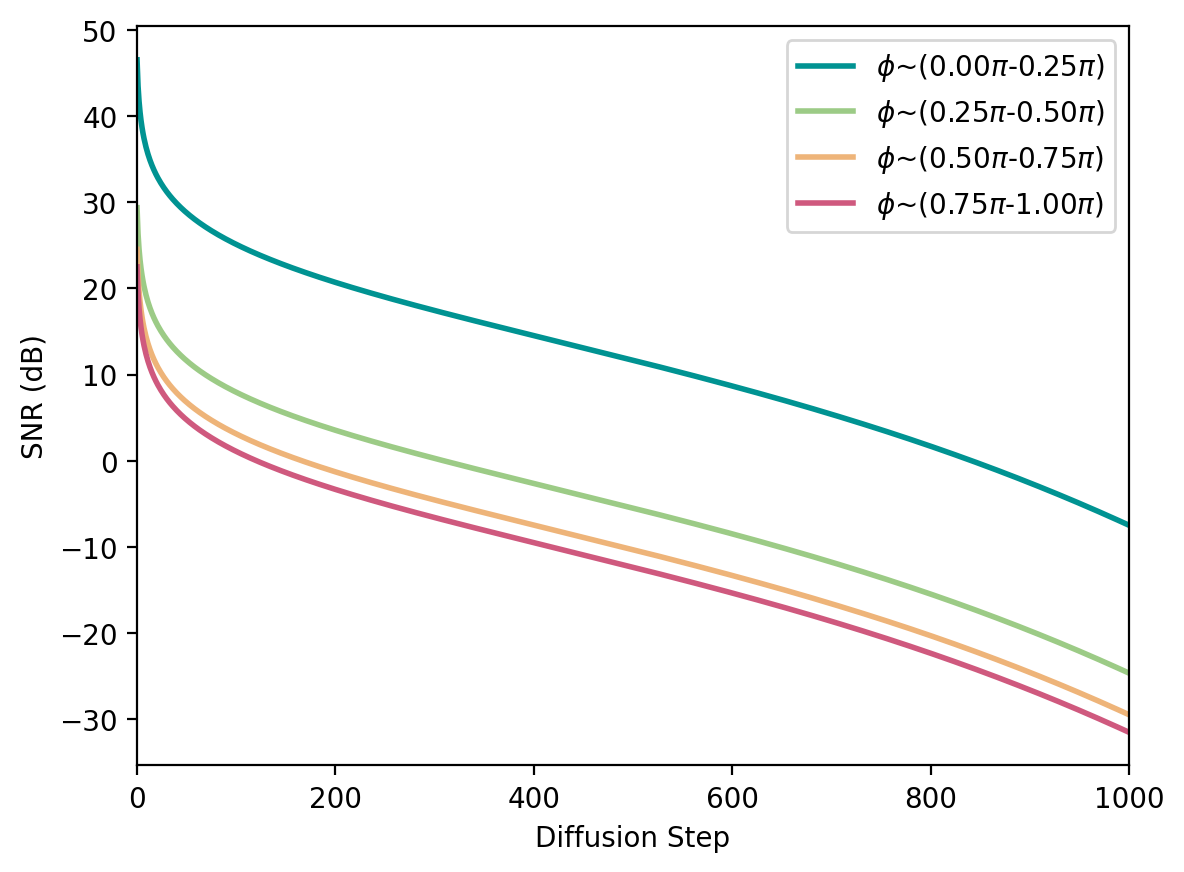}
        \caption{
        \textbf{Signal-to-Noise Ratio (SNR) of different frequency bands at the forward diffusion process.} Each curve corresponds to a spatio-temporal frequency band of the latent code $z_t$ when adding noise at training. The pattern indicates a much slower corruption on low-frequency components.
        }
        \label{fig:snr_change}
    \end{minipage}
    \hfill
    \centering
    \begin{minipage}[t]{0.47\textwidth}
        \centering
        \includegraphics[width=1\textwidth]{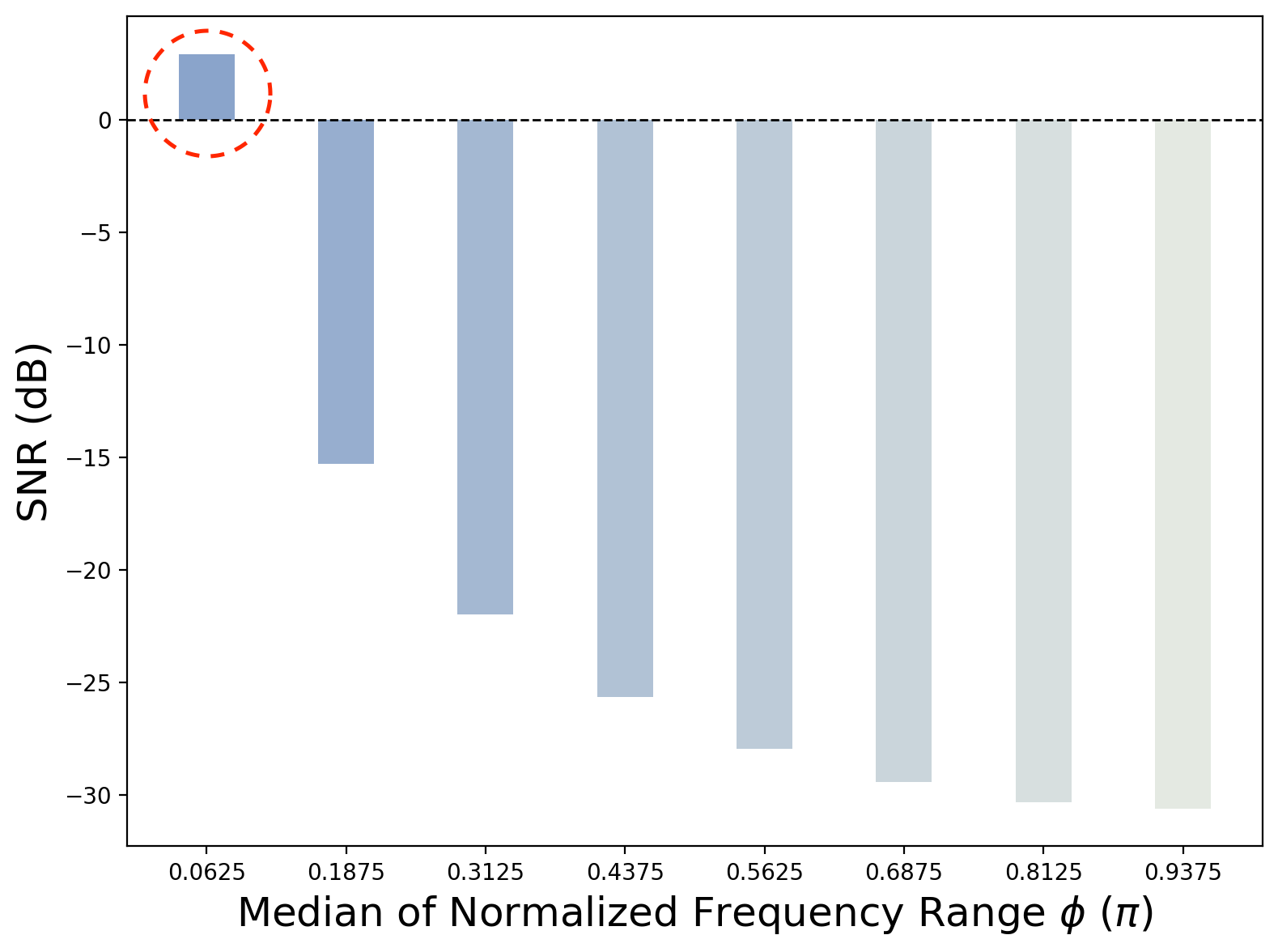}
        \caption{
        \textbf{Frequency Distribution of the SNR in the initial noise.} When training with the typical Stable Diffusion Noise Schedule, the SNR of the initial noise is extremely high in low-frequency components, even larger than 0 dB (red circle). This indicates a severe information leak at the low-frequency band.
        }
        \label{fig:initial_snr_distribution}
    \end{minipage}
\end{figure}

During \textit{inference}, an initial latent $\hat{z_T}$ is first initialized, typically as a Gaussian noise sampled from normal distribution:
\begin{gather}
    \hat{z_T} = \epsilon' \sim \mathcal{N}(0,\mathbf{I}).
    \label{eq:inference-init}
\end{gather}
Then the trained network $\epsilon_\theta$ is used to iteratively denoise the noisy latent to a clean latent $\hat{z_0}$ through DDIM sampling~\cite{song2020ddim}, which is then decoded with decoder $\mathcal{D}$ to obtain video frames $\hat{x_0}$.

\begin{figure}
   \begin{center}
      \includegraphics[width=0.99\linewidth]{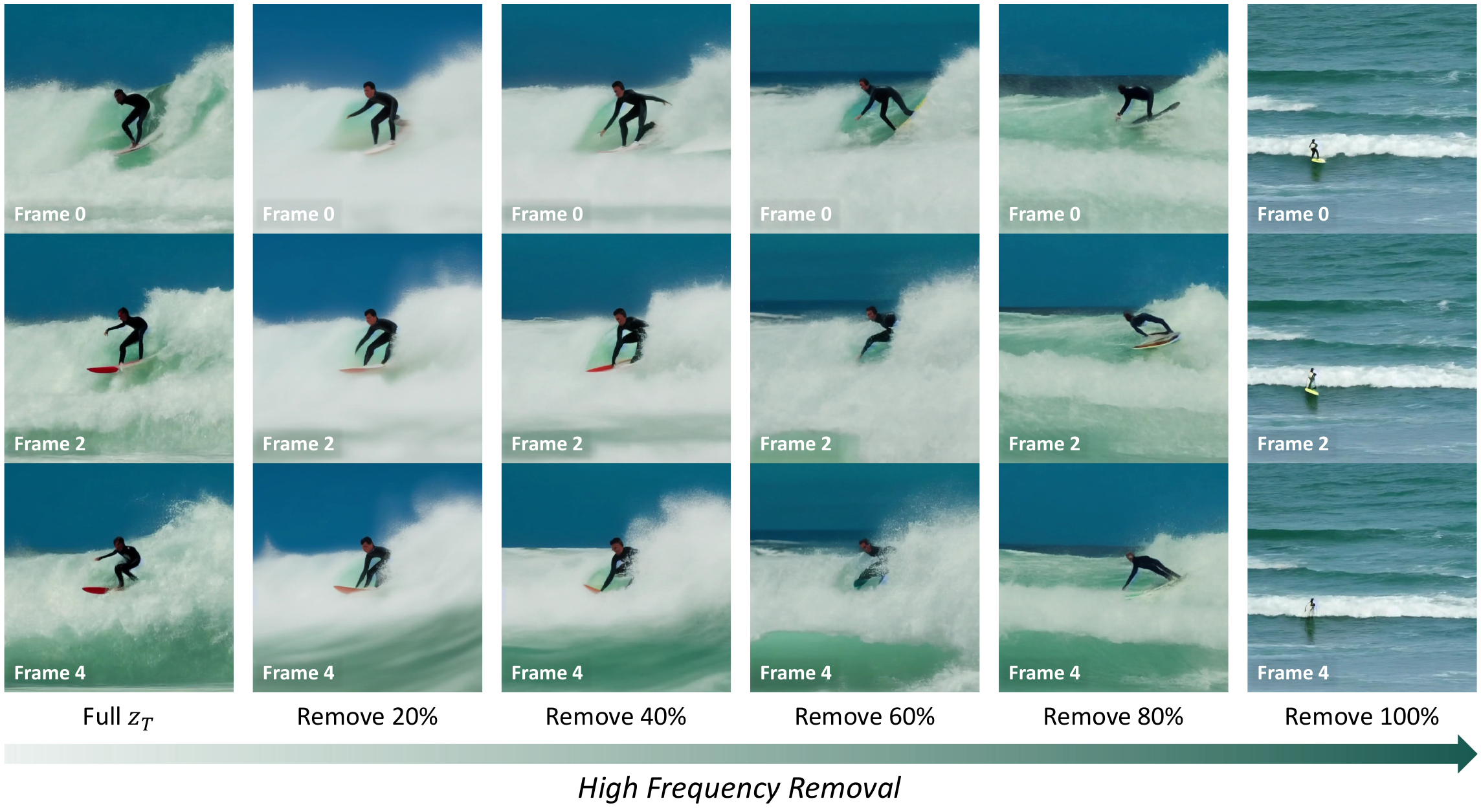}
   \end{center}
   \caption{\textbf{Role of Initial Low-Frequency Components.} 
   Each column shows three frames generated from the mixed initial noise.
   We observe that even if the majority (\eg, 80\%) of high frequencies are replaced, the generated results still remain largely similar to the original ``Full $z_T$'' frames, indicating that the overall distribution of the generated results
   is determined by the low-frequency components of the initial noise.
   }
   \label{fig:influence}
\end{figure}

\subsection{The Initialization Gap}
\label{sec:gap}
\begin{figure}[t]
   \begin{center}
      \includegraphics[width=0.75\linewidth]{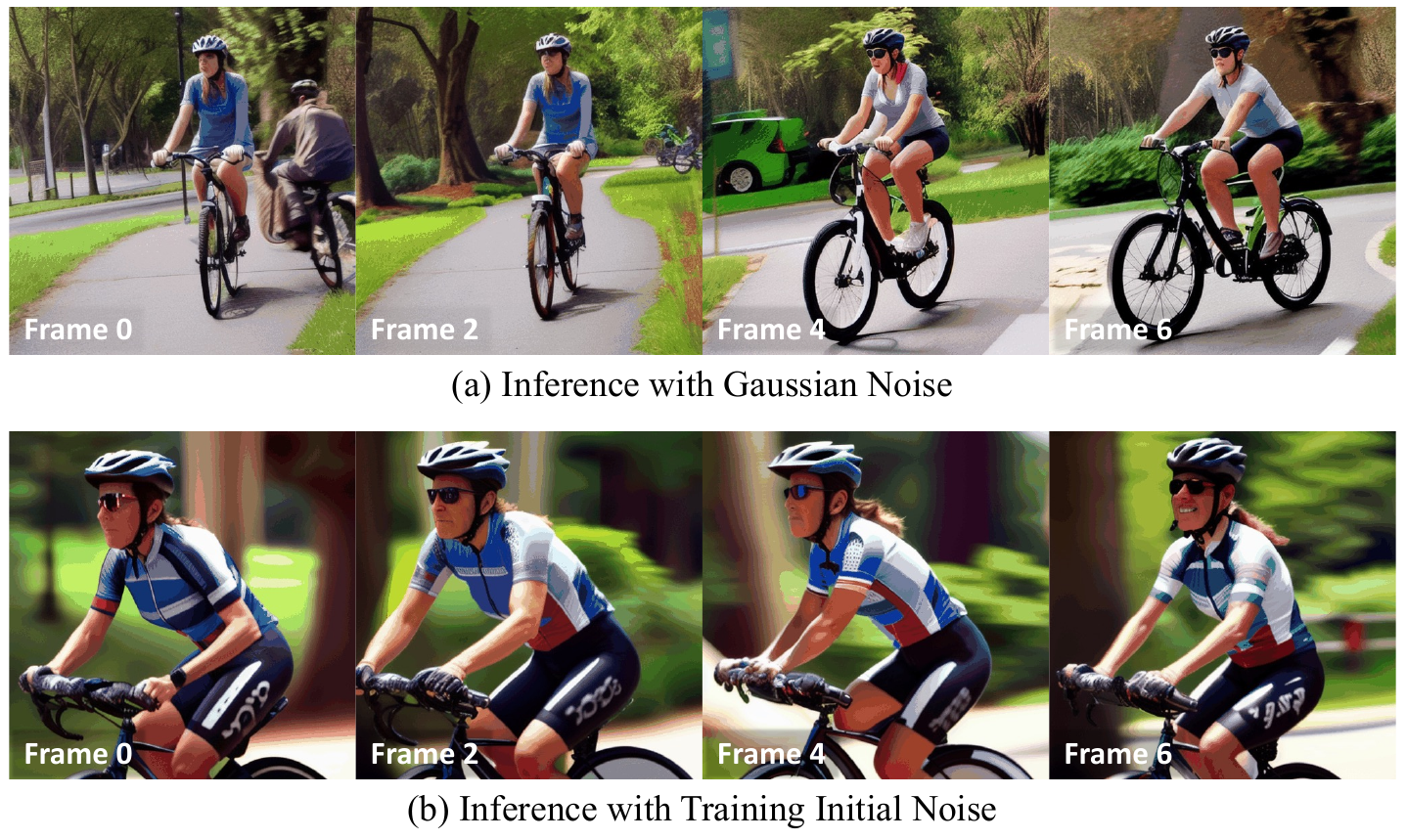}
   \end{center}
   \caption{\textbf{Initialization Gap.} (a) With randomly initialized Gaussian noise for different frames, the sampled video exhibits inconsistency among frames. (b) When we start from noisy latent obtained from the diffusion process from real videos, the generated video is temporally consistent. This is because the initial noise is aligned with training stage and it contains correlated information among different frames in nature.
   }
   \label{fig:naive}
\end{figure}

\noindent
\textbf{Information Leakage at Training.}
At training stage, the network learns to denoise the corrupted latent obtained from the forward diffusion process. However, we find the commonly used diffusion strategy has difficulty in fully corrupting information from real videos, especially in their spatio-temporal low-frequency band. To better demonstrate this phenomenon, we utilize Signal-to-Noise Ratio (SNR) to measure the amount of preserved information at the forward diffusion process. \cref{fig:snr_change} shows the SNR measurements of the noisy latent $z_t$ (as defined in \cref{eq:training-init}) corrupted from a random video clip using Stable Diffusion noise schedule. The figure reveals an obvious pattern wherein the low-frequency components (blue-green curve) exhibit a significantly slower corruption rate compared to the high-frequency components (red curve), which aligns with our observation in \cref{fig:decomposition}. Furthermore, we analyze the average SNR distribution of the initial noises $z_T$ ($T$=$1000$) on UCF-101, and find that the SNR in low-frequency band is even larger than 0 dB, indicating a severe leakage of low-frequency information into the initial noise (\cref{fig:initial_snr_distribution}).

These observations demonstrate the existence of an implicit gap between the training and inference processes. Specifically, the noise introduced during training is insufficient to completely corrupt video information, causing the low-frequency components of the initial noise (\ie, latent at $t$=$1000$) persistently contain spatio-temporal correlations. However, during the inference process, the video generation model is tasked with generating coherent frames from non-correlated Gaussian noise. This presents a considerable challenge for the denoising network, as its initial noise lacks spatio-temporal correlations at inference. For instance, as illustrated in \cref{fig:naive}, the ``biking'' video generated from Gaussian noise exhibits unsatisfactory temporal consistency. In contrast, when using the corrupted latent obtained through the forward diffusion process from real videos as initial noise, the generated frames showcase improved temporal consistency. 

\begin{figure}[t]
   \begin{center}
      \includegraphics[width=1.0\linewidth]{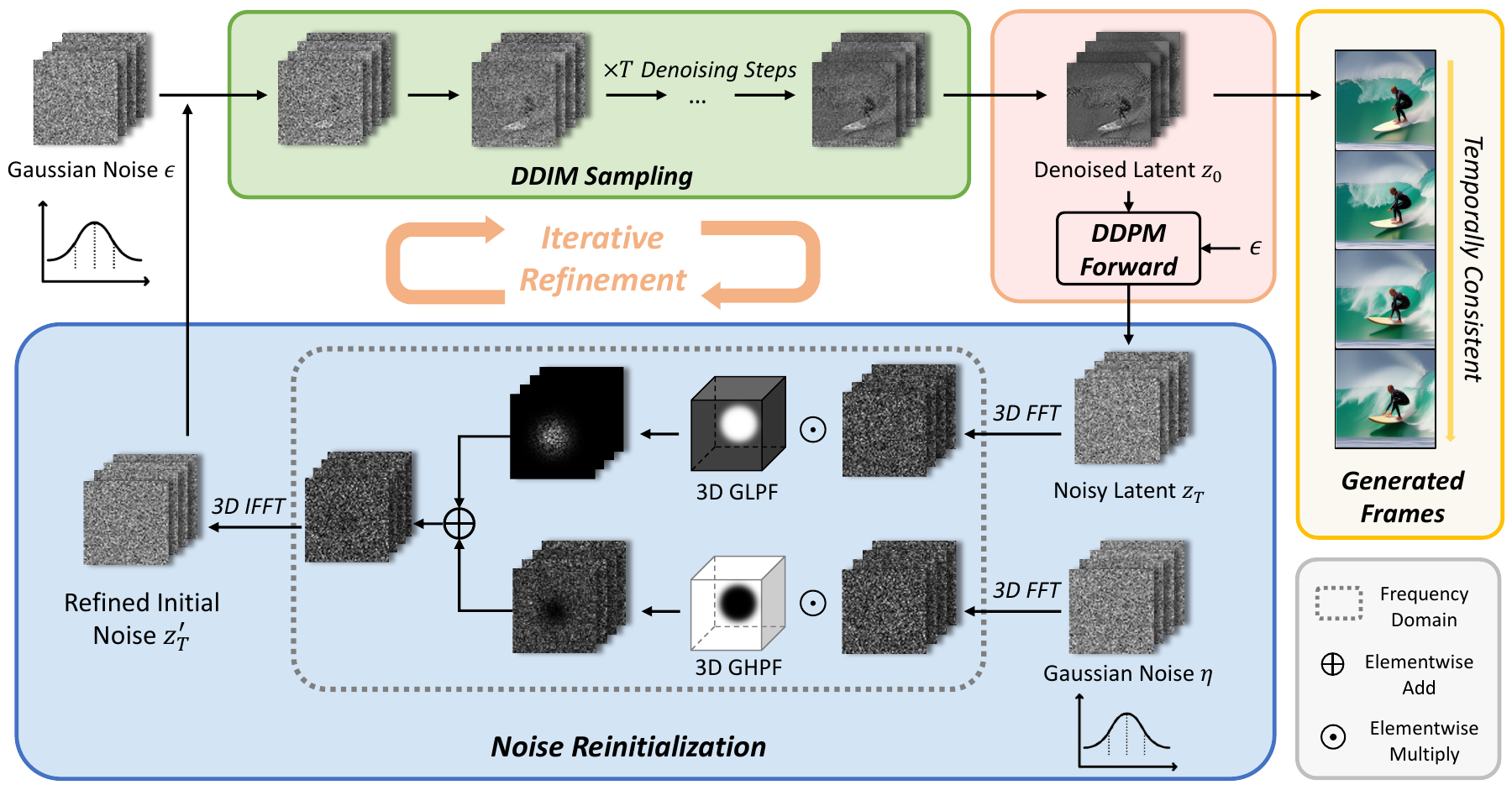}
   \end{center}
   \caption{\textbf{Framework of FreeInit}. 
   FreeInit refines the low-frequency components of the initial noise in an iterative manner. During inference, a Gaussian Noise is first initialized and goes through the standard DDIM sampling process. The resulting denoised latent $z_0$ is then diffused using the original Gaussian Noise $\epsilon$, through DDPM forward process. With the obtained noisy latent $z_T$ which contains richer low-frequency information, a noise reinitialization process is further performed: $z_T$ is firstly transformed into frequency domain through 3D FFT, then its spatio-temporal low-frequency components are fused with the high-frequency from a randomly sampled Gaussian noise $\eta$, bringing flexibility for refinement in higher frequency band. After transforming back to time domain, the refined $z'_T$ is used as the initial noise for the next iteration.
   }
   \label{fig:method}
\end{figure}

\noindent\textbf{Influence of Initial Low-frequency Components.} 
Considering the SNR gaps of initial noise between training and inference, we further investigate the influence of the low-frequency components of initial noise. 
A noisy latent $z_T$ is first obtained from diffusing a real video. 
Then its high-frequency components are gradually removed and replaced with that of a random Gaussian noise, only keeping low-frequencies unchanged. 
Finally, the mixed latent is used as initial noise for inference. As shown in \cref{fig:influence}, it is evident that variations in high-frequency band have a negligible impact on the overall generation results. Remarkably, the overall distribution of the generated outcomes remains stable, even when employing only 20\% of the original initial latent information from the low-frequency band. When all information is removed, the denoising process equates with pure Gaussian noise initialization, which leads to relatively poor generation results. This observation highlights two key conclusions: 1) the low-frequency components of the initial noise play a dominant role at inference, and 2) the quality of low-frequency components is crucial for the generation quality. Our hypothesis is that this is due to the aforementioned information leak during training, which biases the denoising towards the low-frequency components of initial noise. These conclusions motivate us to propose a concise yet effective strategy for enhancing the inference quality of video diffusion models.

\section{FreeInit}
\label{sec:approach}

Motivated by the above analysis, we propose a method for relieving this gap by progressively refining the low-frequency components of the initial noise using the inherent power of the diffusion model. We refer to this method as \textbf{\textit{FreeInit}}, which substantially improves the generation quality without additional training or fine-tuning. The pipeline is illustrated in \cref{fig:method}.

\noindent
\textbf{Denoise and Diffuse.}
During the inference process, an independent Gaussian noise $\epsilon$ is first initialized,  which then undergoes the DDIM sampling process to yield a primary denoised latent $z_0$. Subsequently, we obtain the noisy latent $z_T$ of the generated latent $z_0$ through the DDPM forward diffusion process, \ie, adding noise to diffuse $z_0$ to $z_T$. Since $z_T$ still preserves structural information from the denoised $z_0$ due to the information leakage, its low-frequency components have a better spatio-temporal correlation compared to $\epsilon$. It is worth noting that, during this forward diffusion process, we have observed adding randomly sampled Gaussian noise could introduce significant uncertainty in the mid-frequency band, compromising the spatio-temporal correlation. Consequently, we opt to utilize the same original noise $\epsilon$ used in DDIM sampling when diffusing $z_0$ to $z_T$. The mathematical representation of this process is as follows:
\begin{align}
\label{eq:training-init-free}
    z_T &= \sqrt{\overline{\alpha}_T} z_0 + \sqrt{1-\overline{\alpha}_T}\epsilon \\ 
     &= \sqrt{\overline{\alpha}_T} (DDIM_{sample}(\epsilon)) + \sqrt{1-\overline{\alpha}_T}\epsilon , \nonumber
\end{align}
where $\overline{\alpha}_T$ is aligned with the $\beta$ schedule used at training, \eg, Stable Diffusion schedule.
 
\noindent\textbf{Noise Reinitialization.} To maintain alignment with the SNR distribution at training stage, we propose a noise reinitialization strategy, which is essential for the improvement of temporal consistency. Specifically, we employ a spatio-temporal frequency filter to combine the low-frequency components of the noise latent $z_T$ with the high-frequency components of a random Gaussian noise $\eta$, resulting in a reinitialized noisy latent $z'_T$. This approach allows us to preserve essential information contained in $z_T$ while introducing sufficient randomness in high-frequency to enhance visual details, complementing its improved low-frequency components. The mathematical operations are performed as follows:
\begin{align}
    \mathcal{F}^{L}_{z_T} &= \mathcal{FFT}_{3D}(z_T) \odot \mathcal{H}, \\
    \mathcal{F}^{H}_{\eta} &= \mathcal{FFT}_{3D}(\eta) \odot (1-\mathcal{H}) , \\
    z'_T &= \mathcal{IFFT}_{3D}(\mathcal{F}^{L}_{z_T} + \mathcal{F}^{H}_{\eta}) ,
\end{align}
where $\mathcal{FFT}_{3D}$ is the Fast Fourier Transformation operated on both spatial and temporal dimensions, $\mathcal{H}$ is a spatial-temporal Low Pass Filter (LPF), $\mathcal{IFFT}_{3D}$ is the Inverse Fast Fourier Transformation. 
 
Finally, this reinitialized noise $z'_T$ serves as the starting point for a new round of DDIM sampling, facilitating the generation of frames with enhanced temporal consistency and visual appearance.

\noindent\textbf{Iterative Refinement of Initial Noise.} 
It is important to note that the aforementioned operations can be iteratively applied. At each iteration, the latent code undergoes improvements in spatio-temporal consistency by refining and preserving the low-frequency information from denoising. After that, it gains flexibility in the high-frequency domain through reinitialization, resulting in an improved initial noise for the subsequent iteration. In this iterative manner, the quality of the initial noise is progressively refined, effectively bridging the distribution gap between training and inference. Ultimately, this iterative process contributes to the overall enhancement of generation quality.

\section{Experiments}
\label{sec:exp}

\subsection{Implementation Details}
To evaluate the effectiveness and generalization of our proposed method, we apply the FreeInit strategy to three publically available diffusion-based text-to-video models: AnimateDiff~\cite{guo2023animatediff}, ModelScope~\cite{wang2023modelscope} and VideoCrafter~\cite{chen2023videocrafter1}. Following~\cite{singer2022makeavideo, ge2023pyoco}, we evaluate the inference performance with prompts from UCF-101~\cite{soomro2012ucf101} and MSR-VTT~\cite{xu2016msr} dataset. For UCF-101, we use the same prompt list as proposed in~\cite{ge2023pyoco}. For MSR-VTT, we randomly sample 100 prompts from the test set for evaluation. We also incorporate diverse prompts from~\cite{huang2023vbench} for qualitative evaluations.

During inference, the parameters of frequency filter for each model are kept the same for fair comparison. Specifically, we use a Gaussian Low Pass Filter (GLPF) $\mathcal{H_G}$ with a normalized spatio-temporal stop frequency of $D_0=0.25$. For each prompt, we first adopt the default inference settings of each model for a single inference pass, then apply 4 extra FreeInit iterations and evaluate the progress of generation quality. All FreeInit metrics in Quantitative Comparisons are computed at the $4^{th}$ iteration.

\subsection{Evaluation Metrics}
\noindent\textbf{Temporal Consistency.} 
To measure the temporal consistency of the generated video, we compute frame-wise similarity between the first frame and all succeeding $N-1$ frames. Noteworthily, one typical failure case in current video diffusion models is semantically close but visually inconsistent generation results. For example in \cref{fig:naive} (a), all frames are semantically aligned (``biking''), but the appearance of the subject and background exhibits unsatisfactory consistency. Consequently, semantic-based features like CLIP~\cite{radford2021clip} are not appropriate for evaluating the visual temporal consistency in video generation. Following previous studies~\cite{ruiz2022dreambooth}, we utilize ViT-S/16 DINO~\cite{caron2021dino, oquab2023dinov2} to measure the visual similarities, denoted as the \textbf{DINO} metric. The metric is averaged on all frames.

\noindent\textbf{Motion Quality.} 
To compensate the possible bias of the temporal consistency metric toward over-smoothed videos, we further provide metrics to evaluate the motion quality of the generated videos: 
\textbf{1) Fréchet Video Distance (FVD).} We follow \cite{ge2023pyoco} to perform zero-shot text-to-video generation on UCF-101 and sample 2,048 videos to compute the FVD between the generated distribution and real distribution. Smaller FVD means the distribution is closer to real videos.
\textbf{2) Motion Smoothness (MS) and Dynamic Degree (DD).} Metrics from VBench~\cite{huang2023vbench} are utilized for further evaluation. We use the generated samples on UCF-101 prompts to compute the scores. The scores of real UCF videos (MS=96.64, DD=76.83) are set as a reference to compute the absolute difference $|\Delta_{UCF}|$. Smaller $|\Delta_{UCF}|$ means the motion quality is more similar to real videos.

\subsection{Quantitative Comparisons}

\begin{table}[t]
  \centering
    \caption{\textbf{Quantitative Comparisons on Temporal Consistency.} FreeInit significantly improves the temporal consistency of all baseline methods.}
    \scriptsize{
        \begin{tabular}{l|cc}
            \Xhline{1pt}
            
                                     & \multicolumn{2}{c}{\textbf{DINO} $\uparrow$}                  \\ \cline{2-3} 
            
            \multirow{-2}{*}{\textbf{Method}} & \multicolumn{1}{c|}{\textbf{UCF-101}}        & \textbf{MSR-VTT}        \\ \Xhline{1pt}
            AnimateDiff~\cite{guo2023animatediff}            & \multicolumn{1}{c|}{85.24}          & 83.24          \\
            AnimateDiff+FreeInit                             & \multicolumn{1}{c|}{\textbf{92.01}} & \textbf{91.86} \\ \hline
            ModelScope~\cite{wang2023modelscope}             & \multicolumn{1}{c|}{88.16}          & 88.95          \\
            ModelScope+FreeInit                              & \multicolumn{1}{c|}{\textbf{91.11}} & \textbf{93.28} \\ \hline
            VideoCrafter~\cite{chen2023videocrafter1}        & \multicolumn{1}{c|}{85.62}          & 84.68          \\
            VideoCrafter+FreeInit                            & \multicolumn{1}{c|}{\textbf{89.27}} & \textbf{88.72} \\ \Xhline{1pt}
        \end{tabular}
  }
  \label{tab:quant_comp_consistency}
\end{table}
\begin{table}[t]
  \centering
    \caption{\textbf{Quantitative Comparisons on Motion Quality.} FreeInit also achieves the best motion quality metrics in most cases.}
\scriptsize{
\begin{tabular}{l|@{\hskip .1in}c@{\hskip .2in}c@{\hskip .2in}c}
\Xhline{1pt}
\textbf{Method}  & \textbf{FVD}$\downarrow$ & \textbf{MS($|\Delta_{UCF}|\downarrow$)} & \textbf{DD($|\Delta_{UCF}|\downarrow$)} \\ \Xhline{1pt}
AnimateDiff~\cite{guo2023animatediff}& 1340.96                          & 89.31 (7.33)                   & 97.03 (20.2)                   \\
AnimateDiff+FreeInit                 & \textbf{1032.47}                 & \textbf{96.60 (0.04)}          & \textbf{75.30 (1.53)}          \\ \hline
ModelScope~\cite{wang2023modelscope} & 785.30                           & 95.00 (1.64)                   & \textbf{80.54 (3.71)}          \\
ModelScope+FreeInit                  & \textbf{702.15}                  & \textbf{96.29 (0.35)}          & 68.61 (8.22)                   \\ \hline
VideoCrafter~\cite{chen2023videocrafter1} & 730.04                           & 90.50 (6.14)                   & 92.62 (15.79)                  \\
VideoCrafter+FreeInit                & \textbf{675.39}                  & \textbf{93.45 (3.19)}          & \textbf{83.27 (6.44)}          \\ \Xhline{1pt}

\end{tabular}
}
\label{tab:quant_comp_quality}
\end{table}

The quantitative comparison results are reported in \cref{tab:quant_comp_consistency,tab:quant_comp_quality}.
According to \cref{tab:quant_comp_consistency}, FreeInit significantly improves the temporal consistency of all base models on both prompt sets, by a large margin from 2.92 to 8.62. 
As for motion quality (shown in \cref{tab:quant_comp_quality}), the FVD metrics of all methods are also remarkably improved by FreeInit, indicating a general enhancement in realism. All MS scores are improved and become closer to realistic videos. Although the dynamic degree is decreased, their differences with the ground truth UCF videos mostly become smaller (\eg, AnimateDiff from 20.2 to 1.53). This proves the generated videos are not over-smoothed, but instead become closer to real video distributions.

We also conduct a User Study to evaluate the results through Temporal Consistency, Text Alignment and Overall Quality, which can be referred to in the Supplementary File.

\subsection{Qualitative Comparisons}
\begin{figure}[t!]
   \begin{center}
      \includegraphics[width=1.0\linewidth]{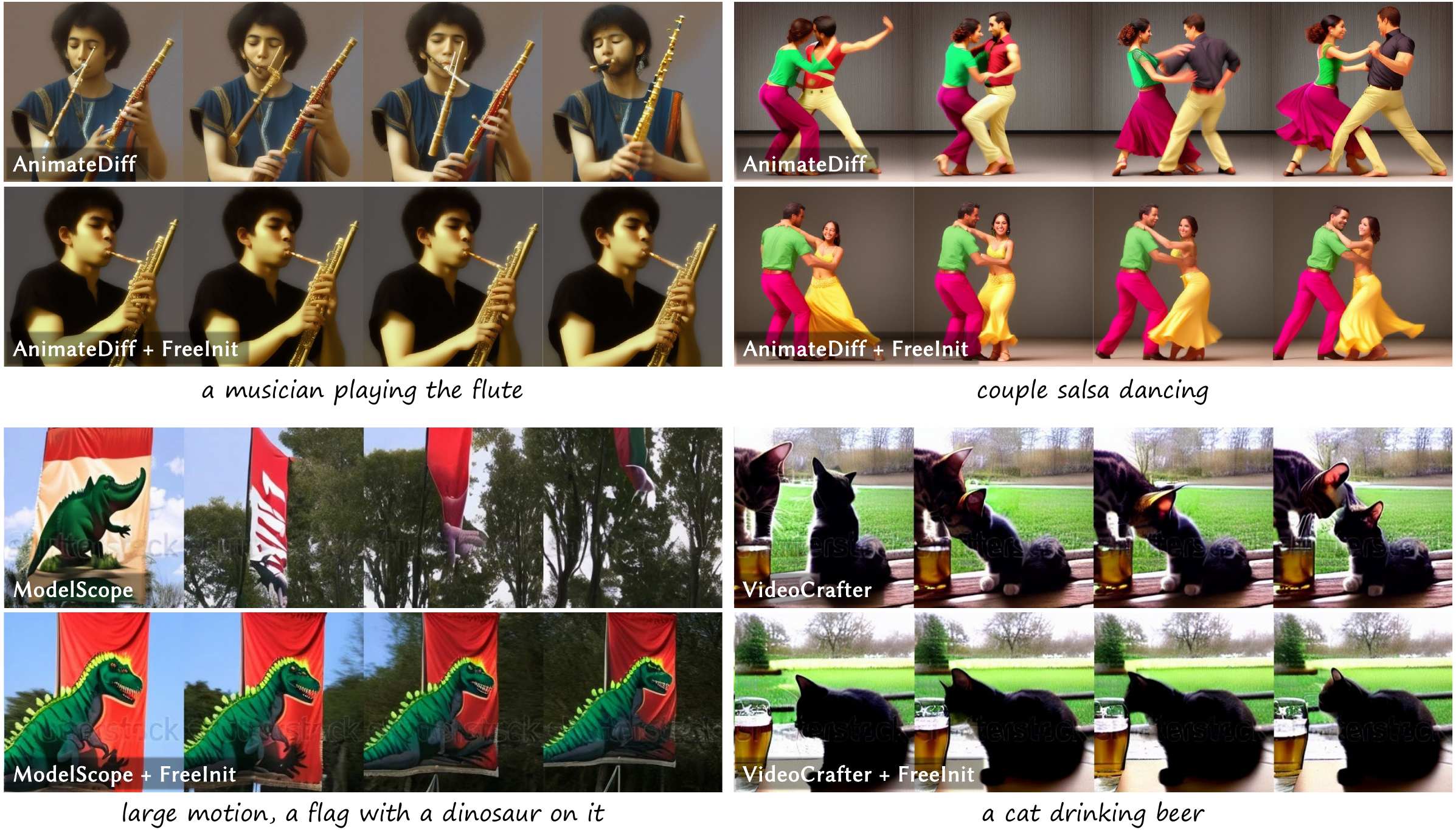}
   \end{center}
      \caption{\textbf{Qualitative Comparisons.} We apply FreeInit to different base models and inference with diverse text prompts. FreeInit significantly improves the temporal consistency and the subject appearance of the generated videos.}
   \label{fig:qualitative}
\end{figure}

Qualitative comparisons are shown in \cref{fig:qualitative}. Our proposed FreeInit significantly improves the temporal consistency as well as visual quality. For example, with text prompt ``a musician playing the flute'', performing FreeInit effectively fixes the temporally unstable artifacts exhibited in vanilla AnimateDiff. More qualitative results are listed in the Supplementary File.

\subsection{Ablation Study}

\begin{table}[t]
  \centering
    \caption{\textbf{Ablation Study on Noise Reinitialization (NR).} Removing NR or changing Gaussian Low Pass Filter (GLPF) to Ideal Low Pass Filter (ILPF) leads to non-optimal results. \textit{ModelName*} refers to \textit{Model+FreeInit}.}
    \scriptsize{
    \begin{tabular}{l|cc}
    \Xhline{1pt}
    \multirow{2}{*}{\textbf{Method}} & \multicolumn{2}{c}{\textbf{DINO} $\uparrow$}                             \\ \cline{2-3} 
                            & \multicolumn{1}{c|}{\textbf{UCF-101}}        & \textbf{MSR-VTT}        \\ \Xhline{1pt}
    AnimateDiff* w/o NR      & \multicolumn{1}{c|}{86.77}          & 85.18          \\
    AnimateDiff* w/ NR-ILPF  & \multicolumn{1}{c|}{87.53}          & 86.17          \\
    AnimateDiff* w/ NR-GLPF  & \multicolumn{1}{c|}{\textbf{92.01}} & \textbf{91.86} \\ \hline
    ModelScope* w/o NR       & \multicolumn{1}{c|}{88.20}          & 90.90          \\
    ModelScope* w/ NR-ILPF   & \multicolumn{1}{c|}{89.04}          & 90.93          \\
    ModelScope* w/ NR-GLPF   & \multicolumn{1}{c|}{\textbf{91.11}} & \textbf{93.28} \\ \hline
    VideoCrafter* w/o NR     & \multicolumn{1}{c|}{86.09}          & 87.11          \\
    VideoCrafter* w/ NR-ILPF & \multicolumn{1}{c|}{87.53}          & 88.01          \\
    VideoCrafter* w/ NR-GLPF & \multicolumn{1}{c|}{\textbf{89.27}} & \textbf{89.33} \\ \Xhline{1pt}
    \end{tabular}
  }
  \label{tab:ablation}
\end{table}

\begin{figure}[t]
    \centering
    \begin{subfigure}{.47\linewidth}
      \centering
      \includegraphics[width=.99\linewidth]{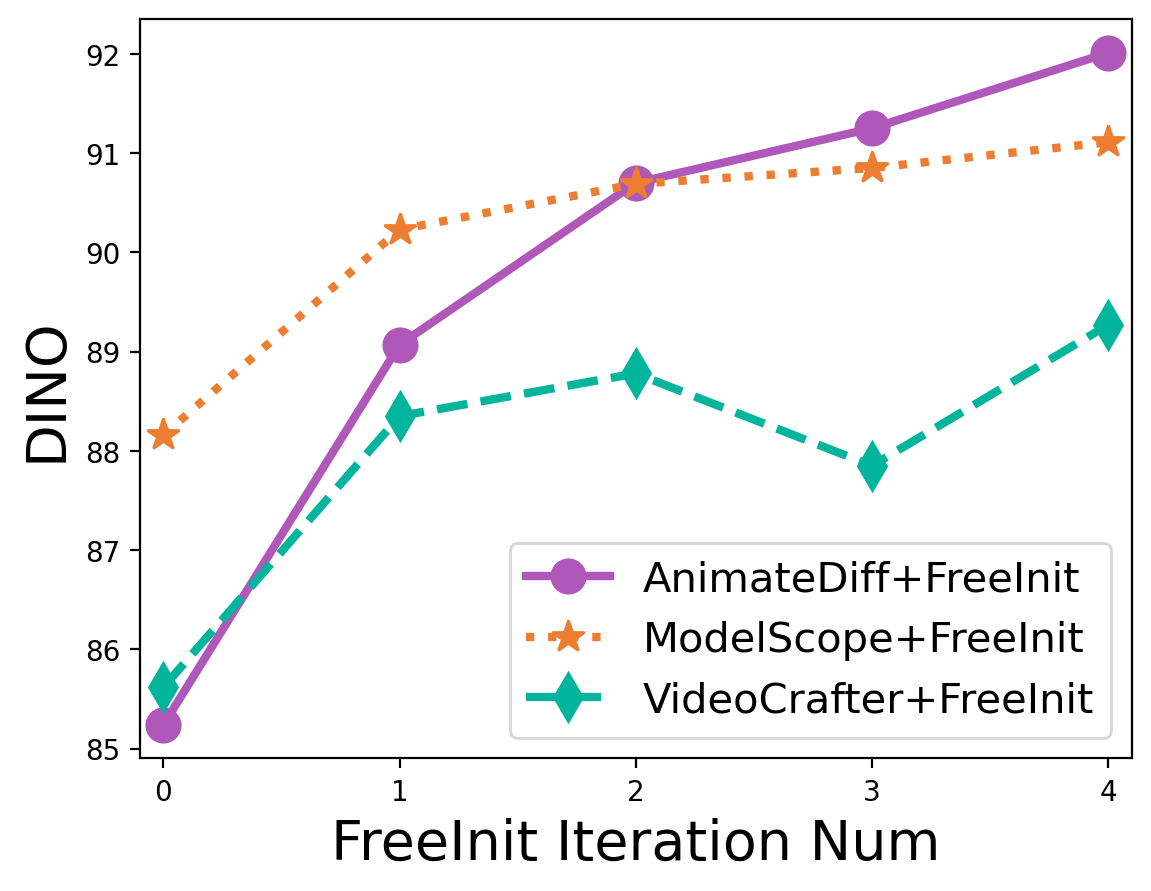}
      \caption{UCF-101}
      \label{fig:sub1}
    \end{subfigure}%
    \hspace{0.05\linewidth}
    \begin{subfigure}{.47\linewidth}
      \centering
      \includegraphics[width=.99\linewidth]{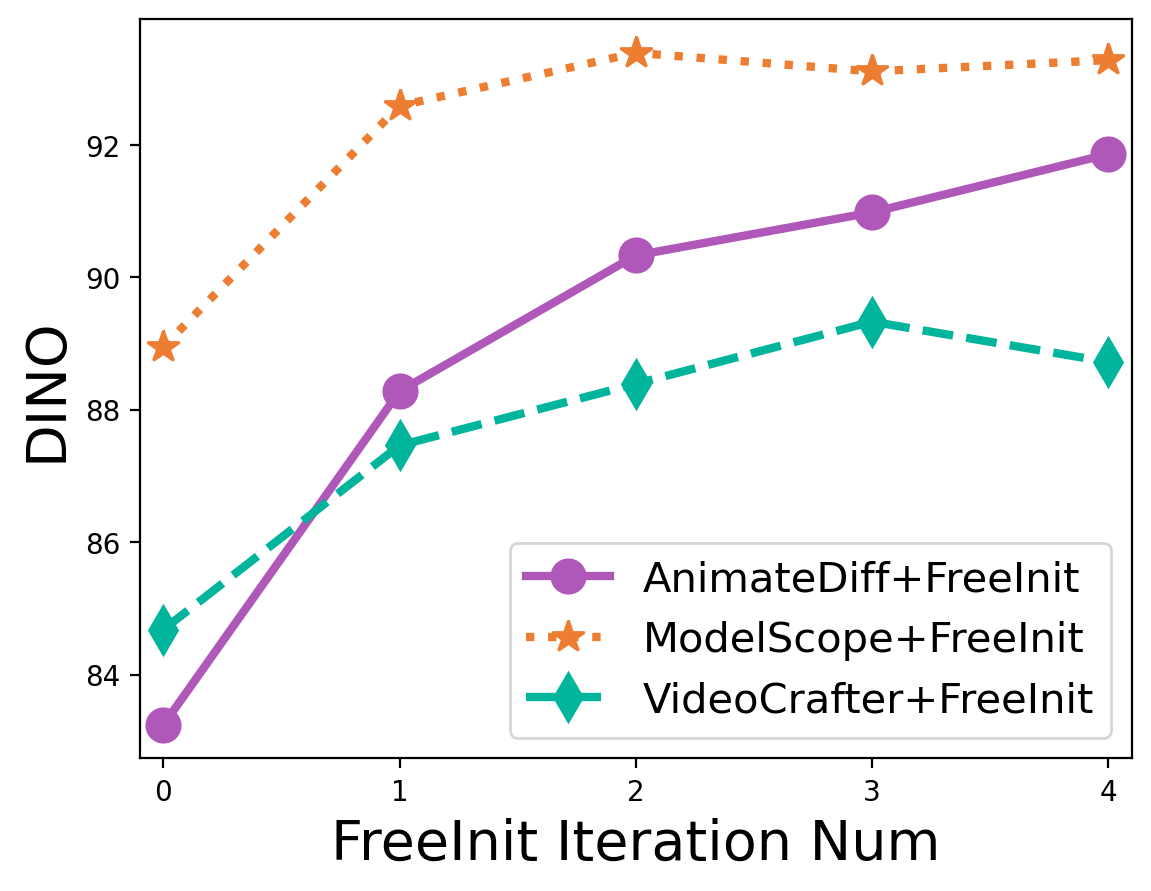}
      \caption{MSR-VTT}
      \label{fig:sub2}
    \end{subfigure}
    \caption{\textbf{Ablation Study on Iteration Number.} We report the DINO scores under different FreeInit iteration numbers on (a) UCF-101 and (b) MSR-VTT.
    More iteration steps mostly lead to better temporal consistency, 
    and the most significant improvement is observed at the $1^{st}$ iteration.
    }
    \label{fig:ablation_iter}
\end{figure}

\begin{figure}[t]
    \centering
    \begin{subfigure}{.47\linewidth}
      \centering
      \includegraphics[width=.99\linewidth]{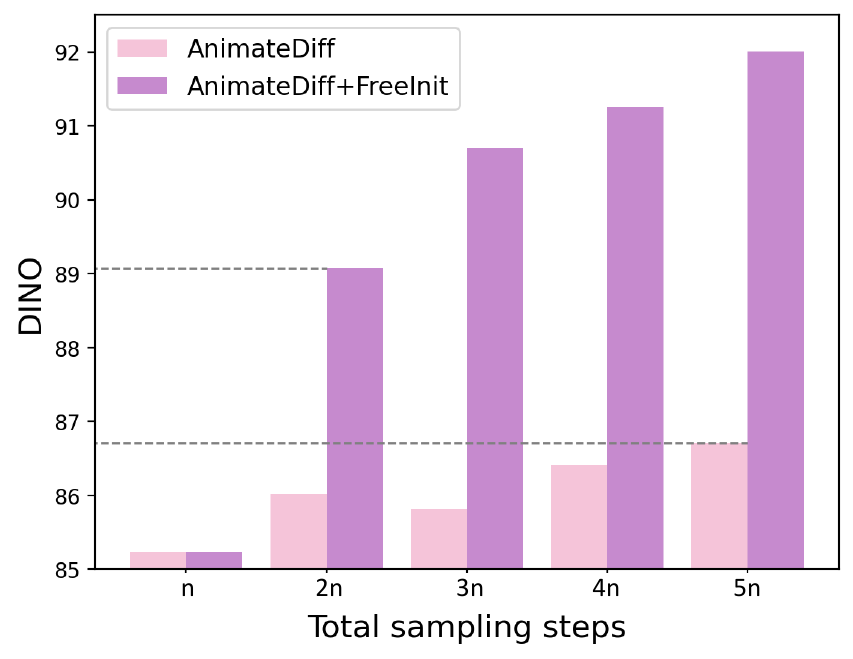}
      \caption{AnimateDiff (n=25)}
    \end{subfigure}%
    \hspace{0.05\linewidth}
    \begin{subfigure}{.47\linewidth}
      \centering
      \includegraphics[width=.99\linewidth]{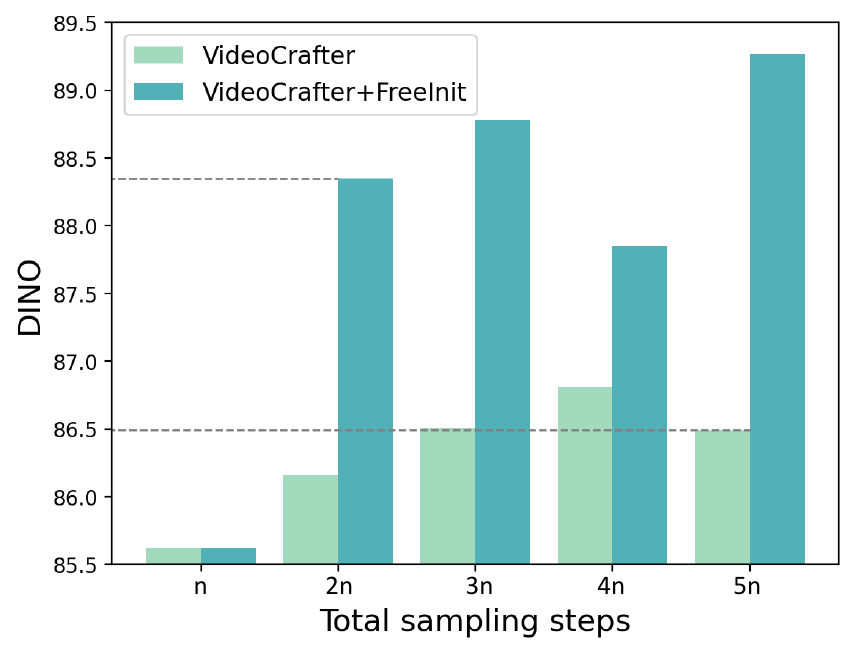}
      \caption{VideoCrafter (n=50)}
    \end{subfigure}
    \caption{\textbf{Comparison with Same Sampling Steps without FreeInit.} We analyze if increasing the DDIM sampling steps for baseline methods would help to improve the temporal consistency on UCF-101. For all base models, the vanilla inference with $5n$ steps is inferior to incorporating FreeInit with $2n$ steps. This indicates that FreeInit is not equivalent to trivially increasing the DDIM sampling steps.}
    \vspace{-10pt}
    \label{fig:same_steps}
\end{figure}

In this section, we quantitatively evaluate the design choices and parameters of FreeInit. Qualitative results can be referred to the Supplementary File.

\noindent\textbf{Influence of Noise Reinitialization and Filter Selection.} To evaluate the importance of Noise Reinitialization in the frequency domain and the choice of filter, we run two FreeInit variants on both datasets with all three base models. Firstly, Noise Reinitialization is totally skipped, \ie, the noisy latent $z_T$ after DDPM Forward Pass is directly used as initial noise for sampling. Secondly, the frequency filter used for Noise Reinitialization is changed from GLPF to ILPF, with the same stop frequency 0.25. The metrics in \cref{tab:ablation} clearly demonstrate that Noise Reinitialization is crucial for improving temporal consistency.
Also, replacing the soft Gaussian filter GLPF with the hard Ideal filter ILPF leads to a performance drop, which reveals the importance of also introducing moderate randomness into mid-frequency and low-frequency components. 
More detailed discussions are in the Supplementary File.

\noindent\textbf{Influence of Iteration Steps.} We show the influence of FreeInit iteration step number in \cref{fig:ablation_iter}. It can be observed that the temporal consistency consistently increases with the iteration step, thanks to the gradually refined initial noise. 
Notably, the largest temporal consistency improvement for each model comes from the \textit{1st} iteration, where FreeInit is applied for the first time. This is because at the \textit{0-th} iteration, the initial noise is non-correlated Gaussian noise, while at the \textit{1st} iteration, low-frequency information is injected into the noise for the first time, largely eliminating the gap between inference noise and training noise.

\subsection{Further Discussion}

\noindent\textbf{Comparison with Same Inference Step without FreeInit.} Since FreeInit uses more than one DDIM sampling pass, it is natural to ask if the quality improvement is due to the increased sampling steps. To answer this question, we compare FreeInit with the typical DDIM sampling strategy using the same total inference steps. As shown in \cref{fig:same_steps}, trivially increasing the DDIM sampling steps only brings little improvement in temporal consistency.
Notably, with just one extra FreeInit iteration (total $2n$ steps), the temporal consistency becomes even better than using $5n$ vanilla DDIM sampling steps that require $\times$2.5 time cost.
This further proves the importance of refining initial noise at inference time: \textit{a good beginning matters more than struggling with a bad initial state}.

\noindent\textbf{Limitations.} As an iterative method, a natural drawback of FreeInit is the increased sampling time. However, incorporating FreeInit leads to much higher performance gain compared to spending more time using the common sampling strategy (\cref{fig:same_steps}).
Furthermore, this issue can be mitigated through a coarse-to-fine sampling strategy. We explain more details and discuss more about the limitations and potential negative societal impacts in the Supplementary File.

\noindent\textbf{Broader Applications.} 
Since the training-inference initialization gap is a common issue, FreeInit is applicable to not only video diffusion models, but also other kinds of diffusion models, \eg, text-to-image models like SDXL~\cite{podell2023sdxl}. Results and discussions are provided in the Supplementary File.

\section{Conclusion}
\label{sec:conclusion}

In this paper, we identify an implicit training-inference gap in the noise initialization of video diffusion models that causes degenerated inference quality: 1) the frequency distribution of the initial noise's SNR is different between training and inference; 2) the denoising process is significantly affected by the low-frequency components of initial noise.
Based on these observations, we propose FreeInit, which improves temporal consistency through the iterative refinement of the spatial-temporal low-frequency component of the initial noise during inference. This narrows the initialization gap between training and inference.
Extensive quantitative and qualitative experiments on various text-to-video models and text prompts demonstrate the effectiveness of our proposed FreeInit.

\section*{Acknowledgement}
This study is supported by the Ministry of Education, Singapore, under its MOE AcRF Tier 2 (MOET2EP20221- 0012), NTU NAP, and under the RIE2020 Industry Alignment Fund – Industry Collaboration Projects (IAF-ICP) Funding Initiative, as well as cash and in-kind contribution from the industry partner(s).

%
%
\bibliographystyle{splncs04}
\bibliography{main}

\clearpage
\section*{Supplementary File}
\appendix

\setcounter{figure}{0}
\setcounter{table}{0}
\setcounter{equation}{0}
\renewcommand\thesection{\Alph{section}}
\renewcommand\thefigure{A\arabic{figure}}
\renewcommand\thetable{A\arabic{table}}
\renewcommand\theequation{A\arabic{equation}}

In this \textit{\textbf{Supplementary File}}, we first explain the SNR distribution computation in \cref{sec:snr}, then tabulate the user study results in \cref{sec:user_study}. More detailed discussions on Noise Reinitialization, filter parameters selection and iteration steps are provided in \cref{sec:more_abl}. Broader applications (\eg. SDXL) are discussed in \cref{sec:sdxl}. We list more implementation details in \cref{sec:implement_detail}. More qualitative comparisons are illustrated in \cref{sec:more_qualitative} to visualize the performance of FreeInit. We further discuss some limitations of FreeInit and its possible social impact in \cref{sec:limitations} and \cref{sec:social}. 
Source code, demo video and more visual comparisons can be found in our project page: \url{https://tianxingwu.github.io/pages/FreeInit/}.

\section{Signal-to-Noise Ratio Distribution}
\label{sec:snr}

In this section, we explain how we derive the frequency SNR distribution of $z_t$ in manuscript Sec. 3.2. 

Mathematically, SNR is defined as the ratio of the power of the signal $P_{signal}$ to the power of the noise $P_{noise}$:
\begin{gather}
    SNR = \frac{P_{signal}}{P_{noise}} = \frac{A^2_{signal}}{A^2_{noise}}
\end{gather}
Where $A$ denotes the amplitude of the signal and the noise.

To measure the frequency distribution of the hidden information in the noisy latent $z_t$ during training, we apply 3D Fourier Transformation $\mathcal{FFT}_{3D}$ to both the clean latent $z_0$ and the Gaussian noise $\epsilon$:
\begin{gather}
    \mathcal{F}_{z_0} = \mathcal{FFT}_{3D}(z_0), \mathcal{F}_{\epsilon} = \mathcal{FFT}_{3D}(\epsilon)
    \label{eq:fz}
\end{gather}

The amplitude spectrum can then be derived with the absolute value of the frequency-domain representation:
\begin{gather}
    \mathcal{A}_{z_0} = |\mathcal{F}_{z_0}|, \mathcal{A}_{\epsilon} = |\mathcal{F}_{\epsilon}|
    \label{eq:az}
\end{gather}

According to Eqn. 4 in manuscript and the linear property of the Fourier transform, the full-band SNR of $z_t$ is derived as:
\begin{gather}
    SNR(z_t)
    = 
    \frac{(\sqrt{\hat{\alpha_t}}A_{z_0})^2}{(\sqrt{1-\hat{\alpha_t}}A_{\epsilon})^2}
    = 
    \frac{\hat{\alpha_t}}{1-\hat{\alpha_t}}
    \frac{A^2_{z_0}}{A^2_{\epsilon}}
    \label{eq:snr}
\end{gather}

Consider a frequency band $\Phi$ with the spatio-temporal frequency range in $\{(f_{s}^{L}, f_{s}^{H}), (f_{t}^{L}, f_{t}^{H})\}$, the SNR of $z_t$ in this frequency band can be calculated using a band-pass filter (BPF), or approximate by summing the amplitudes in the corresponding spatio-temporal range. Converting to logarithm scale, the SNR for frequency band $\Phi$ is finally derived as:
\begin{gather}
    SNR_{dB}(z_t, \Phi)
    =
    10\log_{10}
    \frac{\hat{\alpha_t}}{1-\hat{\alpha_t}}
    \frac{\sum_{f_{s}^{L}}^{f_{s}^{H}} \sum_{f_{t}^{L}}^{f_{t}^{H}} A^2_{z_0}}{\sum_{f_{s}^{L}}^{f_{s}^{H}} \sum_{f_{t}^{L}}^{f_{t}^{H}}A^2_{\epsilon}}
    \label{eq:snrdb}
\end{gather}

\section{User Study}
\label{sec:user_study}

We conduct a User Study to further evaluate the influence of FreeInit. We randomly select 72 diverse text prompts for the test models (VideoCrafter~\cite{chen2023videocrafter1}, ModelScope~\cite{wang2023modelscope} and AnimateDiff~\cite{guo2023animatediff}), and ask 42 participants to vote for the generation results. Specifically, each participant is provided with the text prompt and a pair of synthesized videos, one generated from the vanilla model and the other one with FreeInit. Then the participants vote for the video that they consider superior for Temporal Consistency, Text-Video Alignment and Overall Quality, respectively. The average vote rates are shown in \cref{tab:user_study}. The majority of the votes go to the category using FreeInit under all evaluation metrics, which indicates that FreeInit consistently improves the quality of video generation.

\begin{table}
  \centering
    \caption{\textbf{User Study.} Each participant votes for the image that they consider superior for Temporal Consistency, Text-Video Alignment and Overall Quality, respectively.}
    \footnotesize{
    \begin{tabular}{l|@{\hskip .1in}c@{\hskip .2in}c@{\hskip .2in}c}
    \Xhline{1pt}
    \textbf{Method}       & \makecell[c]{\textbf{Temporal} \\ \textbf{Consistency}} & \makecell[c]{\textbf{Text-Video} \\ \textbf{Alignment}} & \makecell[c]{\textbf{Overall} \\ \textbf{Quality}} \\ \Xhline{1pt}
    VideoCrafter~\cite{chen2023videocrafter1} &      13.10\%                  &         20.24\%         &        18.45\%                             \\
    VideoCrafter+FreeInit                     &      \textbf{86.90\%}                  &         \textbf{79.76\%}         &        \textbf{81.55\%}  \\ \hline \hline
    ModelScope~\cite{wang2023modelscope}      &      19.05\%                  &         23.81\%         &        18.45\%                             \\
    ModelScope+FreeInit                       &      \textbf{80.95\%}                  &         \textbf{76.19\%}         &        \textbf{81.55\%}  \\ \hline \hline
    AnimateDiff~\cite{guo2023animatediff}     &      9.33\%                  &         24.30\%         &        15.97\%                              \\
    AnimateDiff+FreeInit                      &      \textbf{90.67\%}                  &         \textbf{75.70\%}         &        \textbf{84.03\%}  \\ \Xhline{1pt}
    \end{tabular}
  }
  \label{tab:user_study}
\end{table}

\section{More Discussions on Ablation Study}
\label{sec:more_abl}

\begin{figure}[t]
   \begin{center}
      \includegraphics[width=1.0\linewidth]{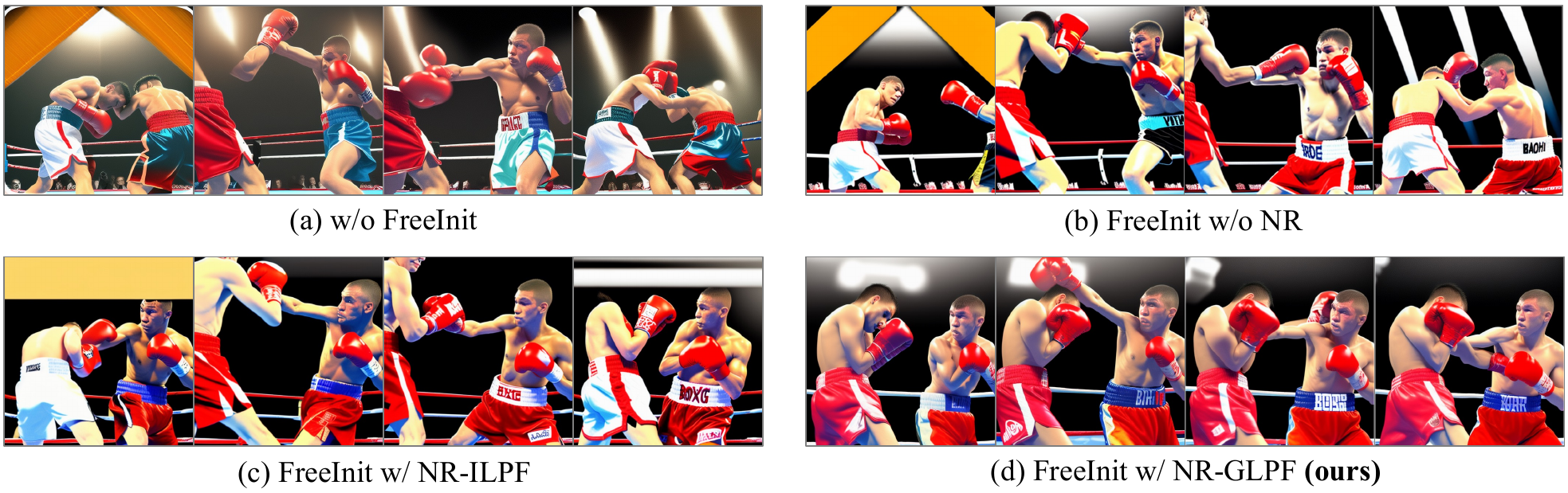}
   \end{center}
   \caption{\textbf{Ablation Study on Noise Reinitialization.} Using Noise Reinitialization with a proper frequency filter is crucial for generating temporally consistent results.
   }
   \label{fig:ablation_filter}
\end{figure}

\noindent\textbf{Influence of Noise Reinitialization and Filter Selection.} The qualitative results generated by the vanilla AnimateDiff model and three ablation variants of FreeInit are illustrated in \cref{fig:ablation_filter}. Aligned with the conclusion in the quantitative results in the manuscript, the visual results indicate that the Noise Reinitialization process is crucial for the refinement of temporal consistency. As depicted in \cref{fig:ablation_filter}, the color of the boxers' clothes and the background suffer from large inconsistencies in the original generation results. Adding the FreeInit refinement loop without Noise Reinitialization still leads to unsatisfactory appearance and undesirable temporal consistency, as no randomness is provided for developing better visual details that complement well with the refined low-frequency components. With Noise Reinitialization performed by Ideal Low Pass Filter (ILPF), the consistency of the subject appearance gains clear improvement, but the large noises (\eg, the yellow area) are still not removed. In comparison, Reinitialization with Gaussian Low Pass Filter (GLPF) is able to remove the large noises, as it introduces an abundant amount of randomness into both mid and high-frequency, creating room for fixing large visual discrepancies. Despite gaining results with better temporal consistency, we also find using Gaussian Low Pass Filter sometimes leads to over-smoothed frames. To alleviate this side effect, the Butterworth Low Pass Filter can be utilized to replace the GLPF for keeping a balance between temporal consistency and visual quality. 

\noindent\textbf{Choice of Cut-off Frequency.} The ablation study on the cut-off frequency of GLPF is shown in \cref{tab:cutoff}. We test different values of $D_0$ with AnimateDiff+FreeInit setting. When $D_0$ increases, the low-frequency component becomes more dominant thus leads to an increase in temporal consistency and motion smoothness. However, large $D_0$ also harms the dynamic degree. We find $D_0=0.25$ achieves an optimal elbow point with the dynamics-consistency trade-off, achieving the most balanced quality. Users may adjust the parameter according to the property of specific base model and their special needs.

\begin{table}[t]
    \centering
    \caption{\textbf{Ablation on Cut-off Frequency.} Our setting achieves a balanced dynamics-consistency trade-off.}
    \footnotesize
        \begin{tabular}{l|ccc}
        \Xhline{1pt}
        \textbf{$D_0$}  & \textbf{DINO}$\uparrow$ & \textbf{MS($|\Delta_{UCF}|\downarrow$)} & \textbf{DD($|\Delta_{UCF}|\downarrow$)} \\ \Xhline{1pt}
        0.125       & 91.54            & 92.83 (3.81)              & 89.52 (12.69)                   \\
        \textbf{0.25 (ours)} & 92.07            & \textbf{96.80 (0.16)}              & \textbf{75.24 (1.59)}                   \\
        0.5         & \textbf{93.91}   & 98.70 (2.06)     & 39.05 (37.78)          \\ \Xhline{1pt}
        
        \end{tabular}
    \label{tab:cutoff}
\end{table}

\begin{figure}[t]
    \centering
    \begin{minipage}[t]{.48\textwidth}
        \centering
        \includegraphics[width=1.0\linewidth]{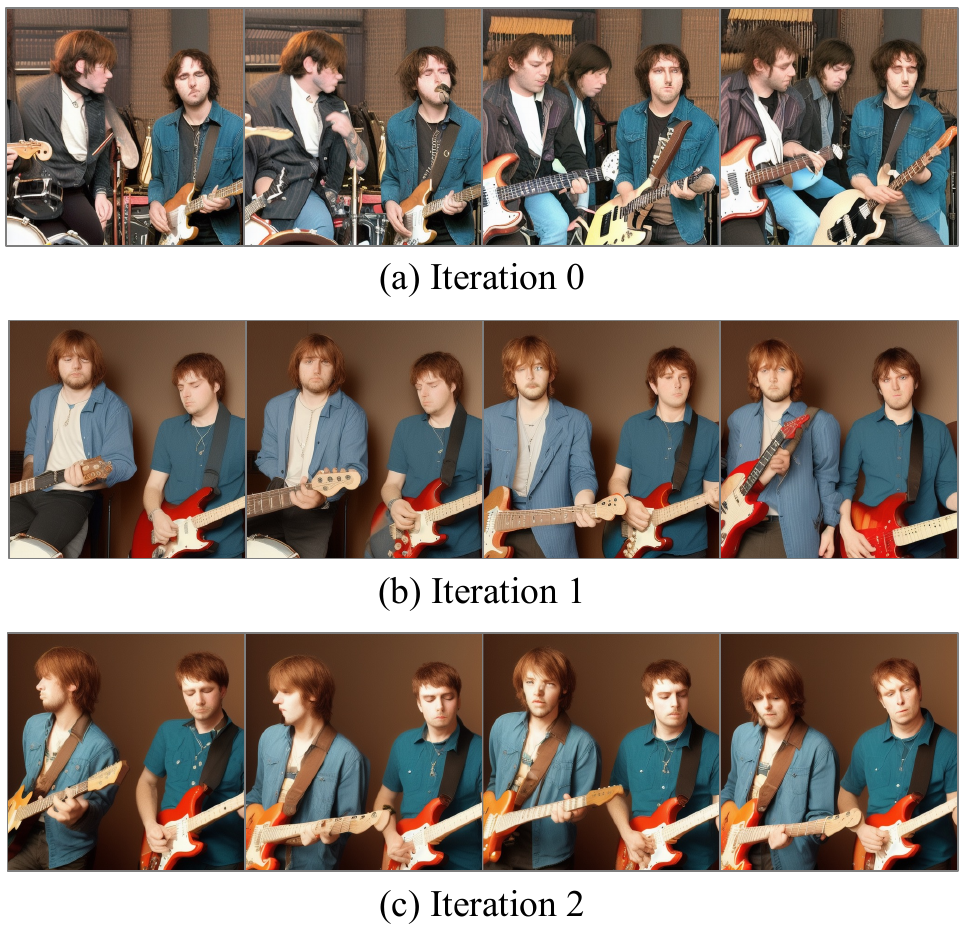}
        \caption{\textbf{Ablation Study on Iteration Steps.} 
        The appearance of the players and instruments becomes more consistent after each iteration. However, it is worth mentioning that 
        the largest consistency leap takes place in the first iteration.
        }
        \label{fig:ablation_steps_1}
    \end{minipage}
    \hfill
    \centering
    \begin{minipage}[t]{0.48\textwidth}
        \centering
        \includegraphics[width=1.0\linewidth]{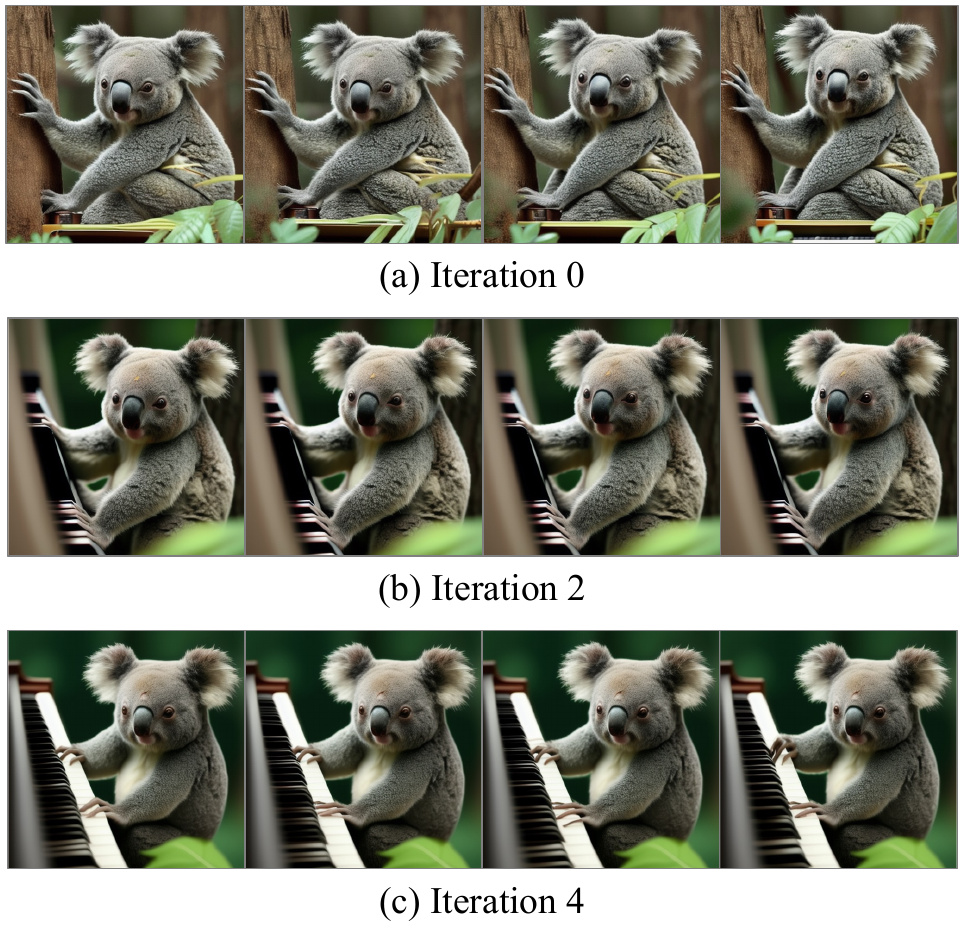}
        \caption{\textbf{Semantic Growth.} The frames are generated with the text prompt ``A koala bear is playing piano in the forest". The missing semantics ``playing piano" gradually grows with the FreeInit iteration.
        }
        \label{fig:ablation_steps_2}
    \end{minipage}
\end{figure}

\noindent\textbf{Influence of Iteration Steps.} Since the low-frequency components of the latent are refined during each FreeInit iteration, more iteration steps normally lead to generation results with better temporal consistency. Nonetheless, the most significant improvement comes from the 1st refinement iteration (iteration 1), as shown in \cref{fig:ablation_steps_1}. We also observe that the missing semantics (\eg, ``playing piano'', as depicted in \cref{fig:ablation_steps_2}) can gradually grow with the FreeInit iteration, thanks to the refined low-frequency components and the randomness introduced by Noise Reinitialization.

\section{Broader Applications}
\label{sec:sdxl}

Since our discovered training-inference initialization gap is a common issue, FreeInit is applicable to not only video diffusion models, but also other kinds of diffusion models, \eg, text-to-image models. 

To enable FreeInit on image models, we remove the temporal dimension of the original spatio-temporal frequency decomposition operation in FreeInit, and use a 2D GLPF with $D_0=0.125$ to implement the Noise Reinitialization.

\begin{figure*}
   \begin{center}
      \includegraphics[width=0.965\linewidth]{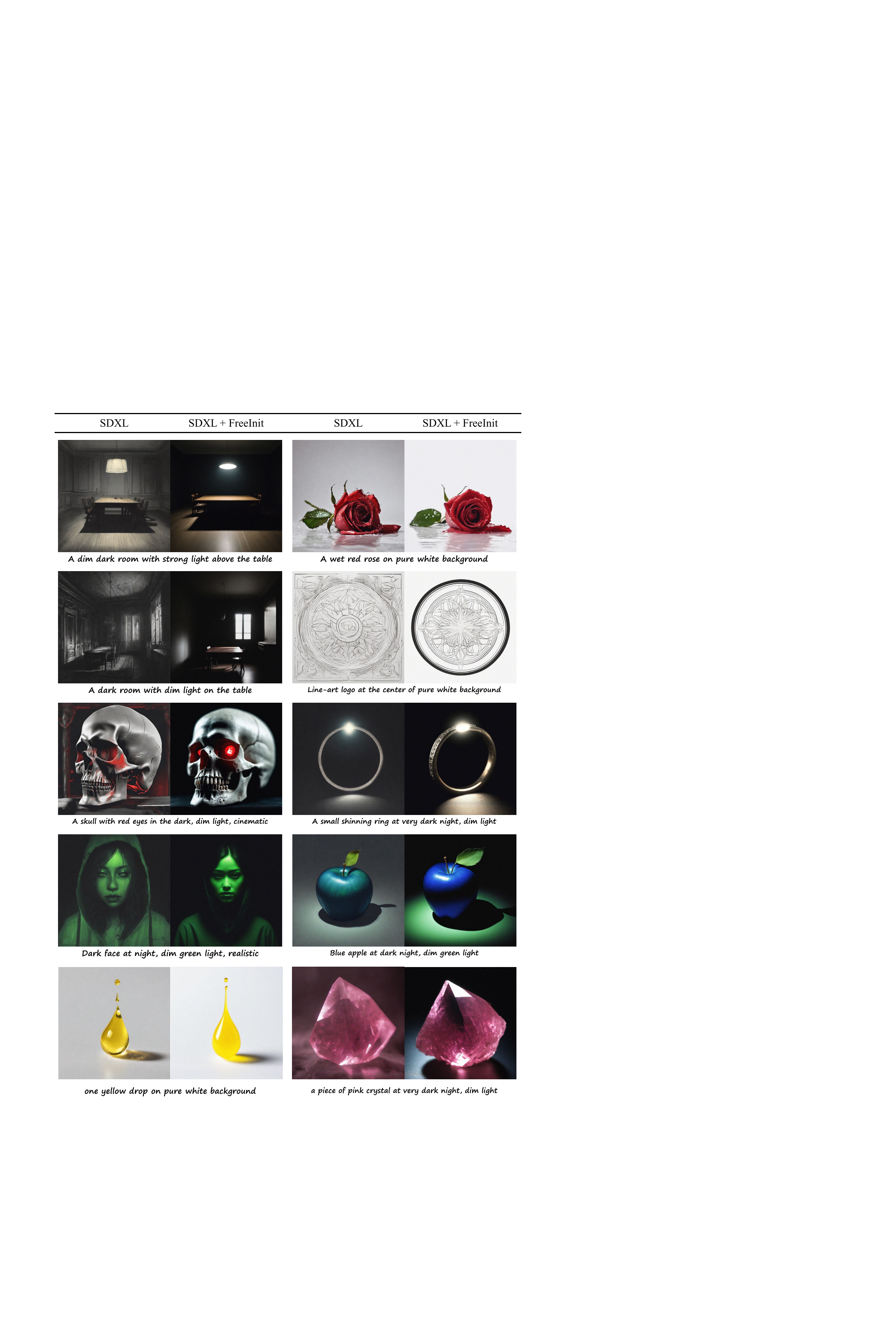}
   \end{center}
   \caption{\textbf{FreeInit for Text-to-Image Diffusion Models.} It enables SDXL~\cite{podell2023sdxl} to generate very dark/bright images with better text alignment and high visual quality.}
   \label{fig:sdxl}
\end{figure*}

We show visual results of adding FreeInit to SDXL~\cite{podell2023sdxl} in \cref{fig:sdxl}. By iteratively refining the initial noise, FreeInit helps SDXL to generate very dark and bright images with improved text alignment and high visual quality. This functionality is similar to the approach proposed in \cite{lin2023common}, while we require no additional training or fine-tuning to narrow the SNR gap.

\section{Implementation Details}
\label{sec:implement_detail}
\noindent\textbf{Base Models.} Three open-sourced text-to-video models are used as the base models for FreeInit evaluation. For VideoCrafter~\cite{chen2023videocrafter1}, the VideoCrafter-v0.9 Base T2V model based on the latent video diffusion models (LVDM)~\cite{he2022lvdm} is adopted. For ModelScope~\cite{wang2023modelscope}, we utilize the \textit{diffusers}~\cite{diffusers} implementation of the text-to-video pipeline, with the text-to-video-ms-1.7b model. For AnimateDiff~\cite{guo2023animatediff}, we use the mm-sd-v14 motion module with the Realistic Vision V5.1 LoRA model for evaluation. 

\noindent\textbf{Inference Details.} Experiments on VideoCrafter and ModelScope are conducted on $256 \times 256$ spatial scale and $16$ frames, while experiments on AnimateDiff are conducted on a video size of $512 \times 512$, $16$ frames. During the inference process, we use classifier-free guidance for all experiments including the comparisons and ablation studies, with a constant guidance weight 7.5. All experiments are conducted on a single Nvidia A100 GPU.

\section{More Qualitative Comparisons}
\label{sec:more_qualitative}
Extra qualitative results on AnimateDiff~\cite{guo2023animatediff}, ModelScope~\cite{wang2023modelscope} and VideoCrafter~\cite{chen2023videocrafter1} are provided in \cref{fig:animatediff_1}-~\ref{fig:videocrafter_2}. 

We also include visual comparisons on the more recent video diffusion model VideoCrafter2~\cite{chen2024videocrafter2} in \cref{fig:videocrafter2}. As can be observed in the results, although VideoCrafter2 generally archives better generation quality, temporal inconsistencies still exist, and FreeInit is able to mitigate the issue.

For more visual results in video format, please refer to our \href{https://tianxingwu.github.io/pages/FreeInit/}{project page}.

\section{Limitations}
\label{sec:limitations}

\noindent\textbf{Inference Time.} Since FreeInit is an iterative method, a natural drawback is the increased sampling time. 
To avoid unnecessary time cost, we only use around 2-3 extra FreeInit iterations in practice, since the largest performance improvement mainly comes from the 1st refinement iteration (\cref{fig:ablation_steps_1}, Manuscript Fig.9), and that the performance increase saturates after 2-3 iterations (Manuscript Fig.9). Along with numerical optimization, we can sample a high-resolution $512\times512\times16$ video with the optimized AnimateDiff+FreeInit in less than 30s on a single A100 GPU.

\begin{figure}[t]
    \centering
    \begin{minipage}[t]{.48\textwidth}
        \centering
        \includegraphics[width=1.0\linewidth]{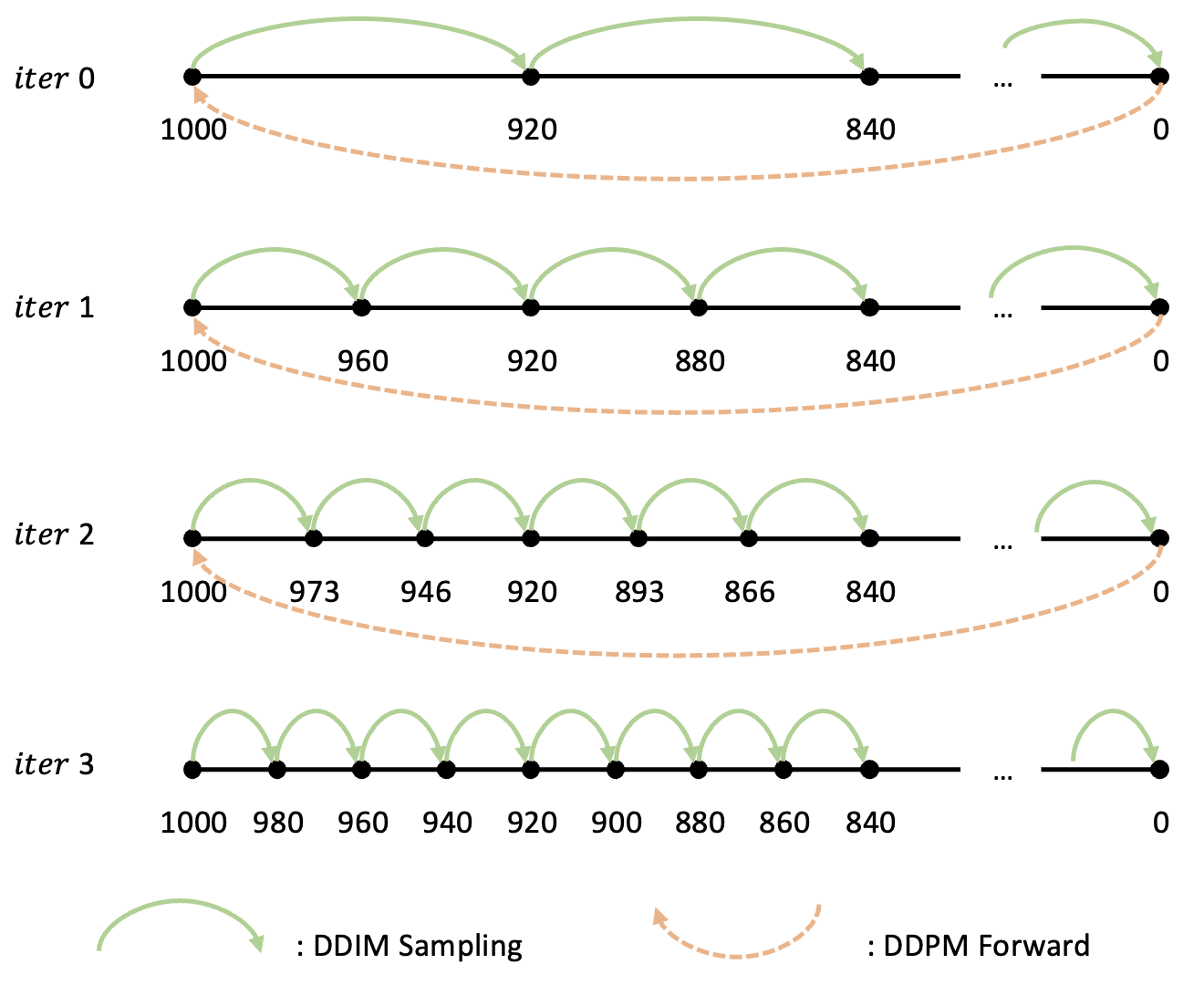}
        \caption{\textbf{Coarse-to-Fine Sampling for Fast Inference.} Early FreeInit iterations use fewer DDIM sampling steps, while latter iterations perform more steps for detailed refinement.
        }
        \label{fig:coarse-to-fine_1}
    \end{minipage}
    \hfill
    \centering
    \begin{minipage}[t]{0.48\textwidth}
        \centering
        \includegraphics[width=1.0\linewidth]{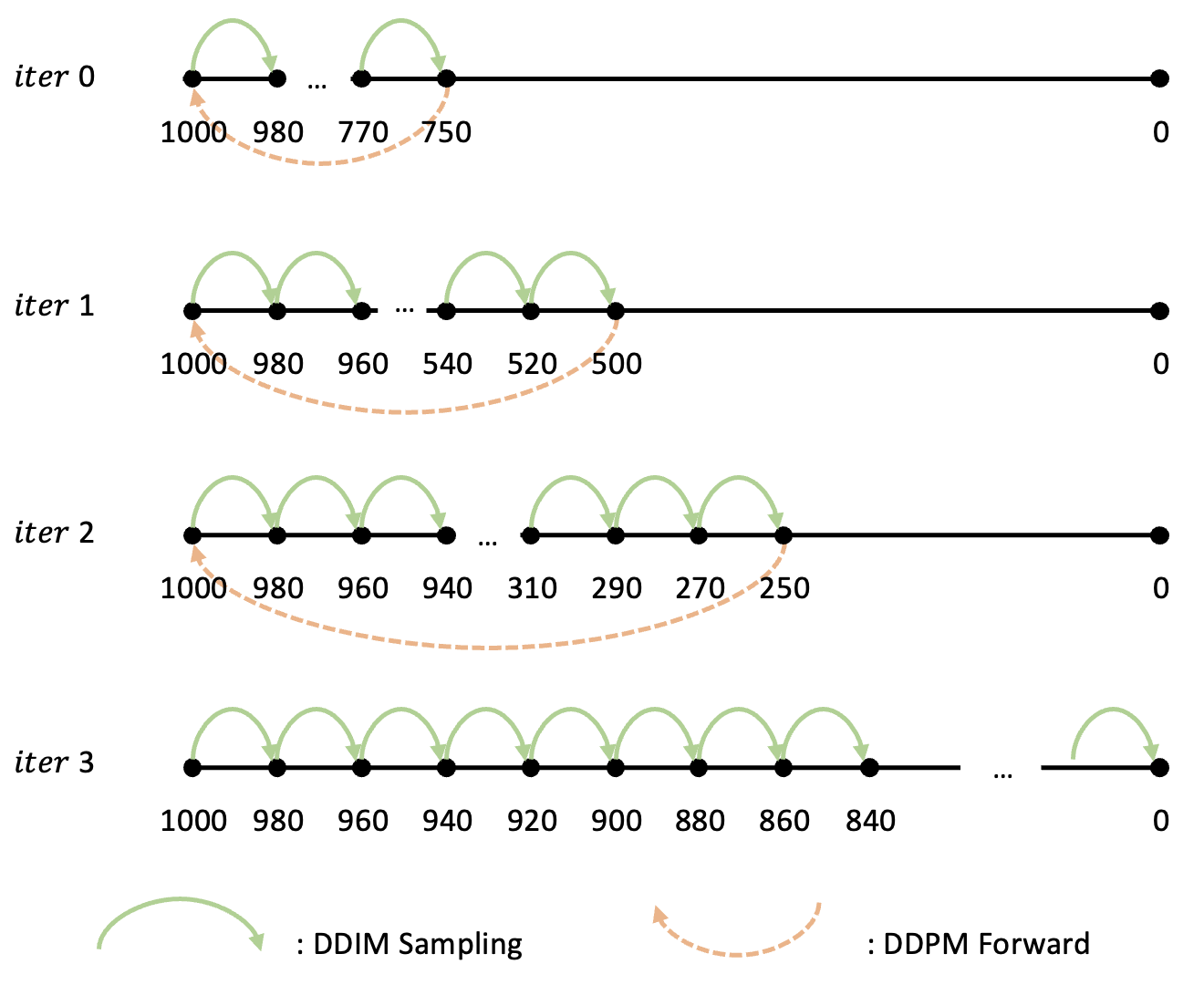}
        \caption{\textbf{Another Design Choice for the Coarse-to-Fine Sampling.} We empirically find that this strategy is sub-optimal compared to the other one in several cases.
        }
        \label{fig:coarse-to-fine_2}
    \end{minipage}
\end{figure}

This issue can be further mitigated through a \textbf{Coarse-to-Fine Sampling} Strategy. Specifically, the DDIM sampling steps of each FreeInit iteration can be reduced according to the iteration step: early FreeInit iterations use fewer DDIM sampling steps for a coarse refinement of the low-frequency components, while the latter iterations perform more steps for detailed refinement. A simple linear scaling strategy can be used for Coarse-to-Fine Sampling:
\begin{gather}
    T_i = [\frac{T}{N} (i + 1)]
\end{gather}
Where $T_i$ is the DDIM steps for iteration $i$, $T$ is the commonly used DDIM steps (\eg, 50), and $N$ is the number of FreeInit iterations. For instance, the fast Sampling process for $T=50$, $N=4$ is illustrated in \cref{fig:coarse-to-fine_1}. Besides, we also test another fast sampling strategy, as illustrated in \cref{fig:coarse-to-fine_2}. The key idea is to keep the step size fixed but apply an early-stop at early iterations. However, we find this strategy is sub-optimal compared to the former one.

\begin{figure}[t]
   \begin{center}
      \includegraphics[width=1.0\linewidth]{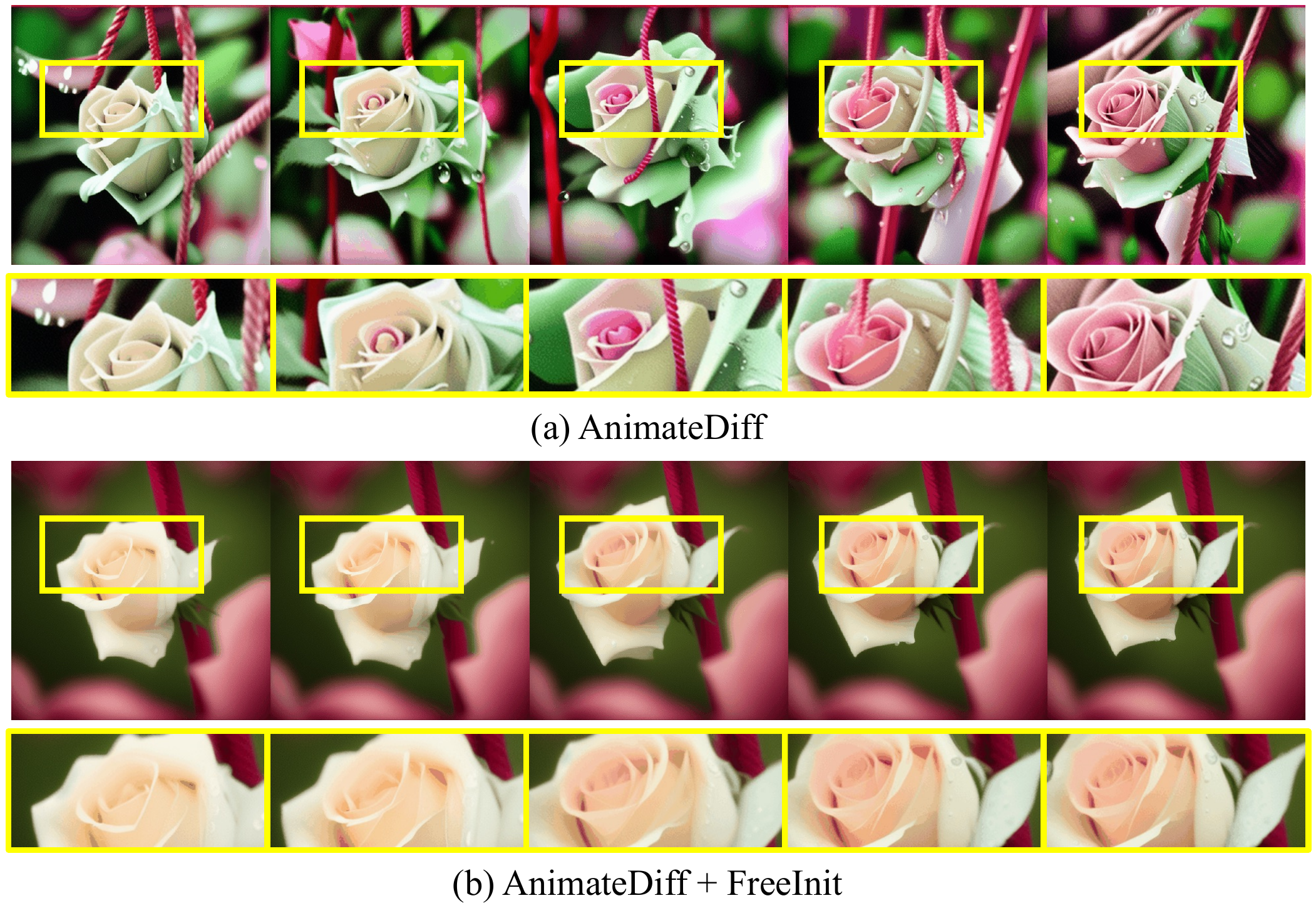}
   \end{center}
   \caption{\textbf{Failure Case.} With input prompt ``a rose swing in the wind with waterdrops'', performing FreeInit improves temporal consistency but falsely removes the fast, small foreground object (waterdrops).
   }
   \label{fig:failure}
\end{figure}

\begin{figure*}[t]
   \begin{center}
      \includegraphics[width=0.99\linewidth]{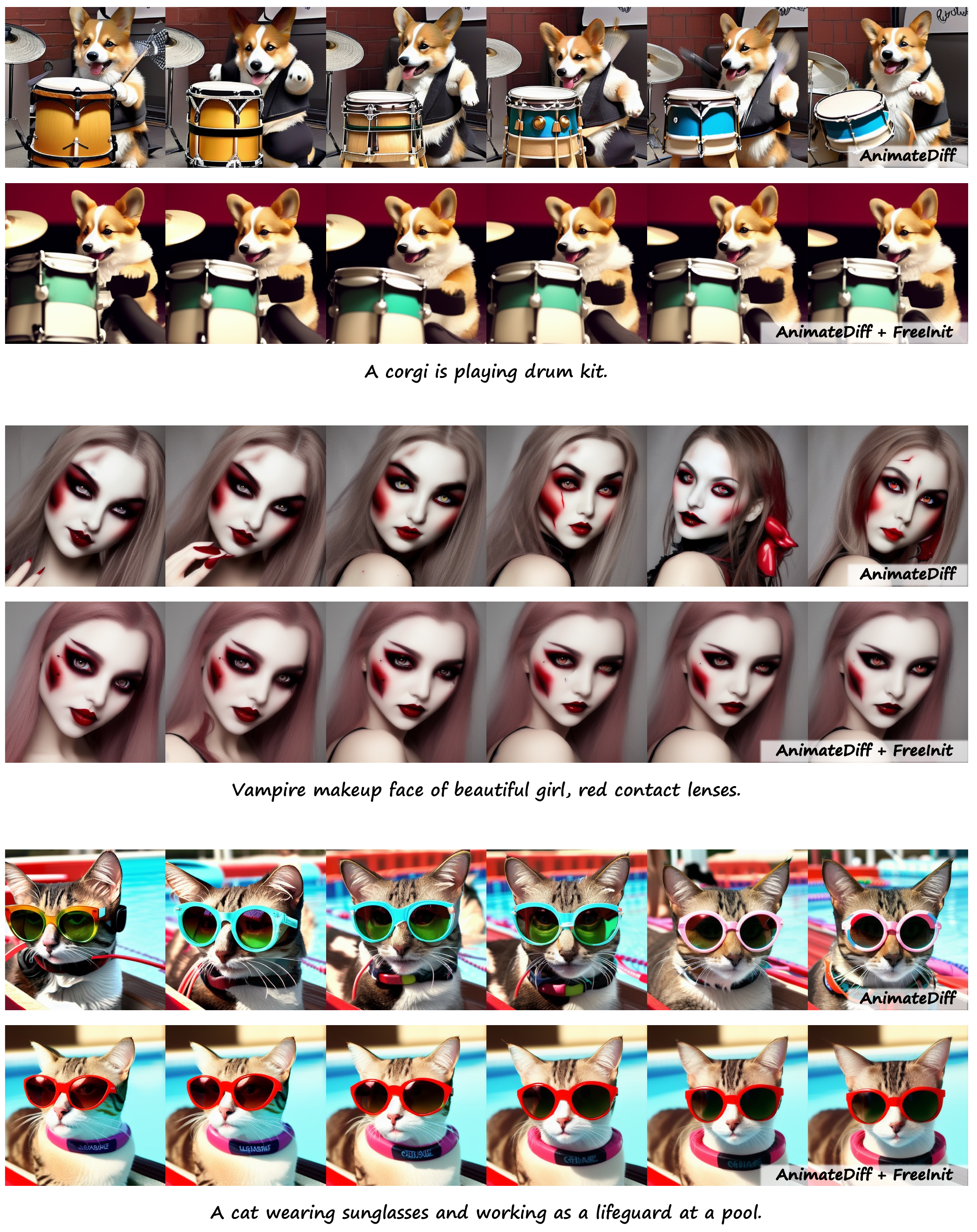}
   \end{center}
   \caption{\textbf{More Qualitative Results of FreeInit on AnimateDiff.}}
   \label{fig:animatediff_1}
\end{figure*}

\begin{figure*}[t]
   \begin{center}
      \includegraphics[width=0.98\linewidth]{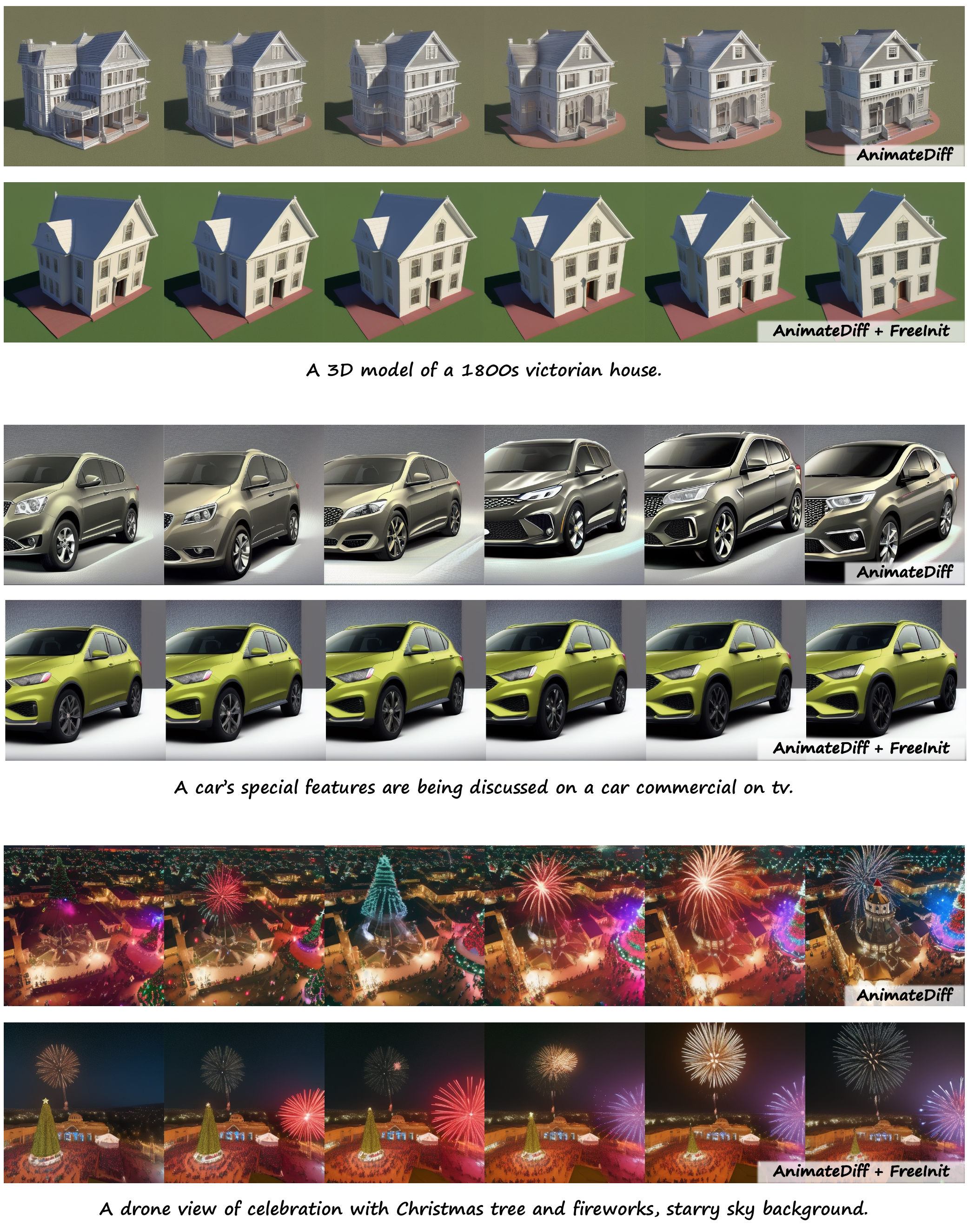}
   \end{center}
   \caption{\textbf{More Qualitative Results of FreeInit on AnimateDiff.}}
   \label{fig:animatediff_2}
\end{figure*}

\begin{figure*}[t]
   \begin{center}
      \includegraphics[width=0.98\linewidth]{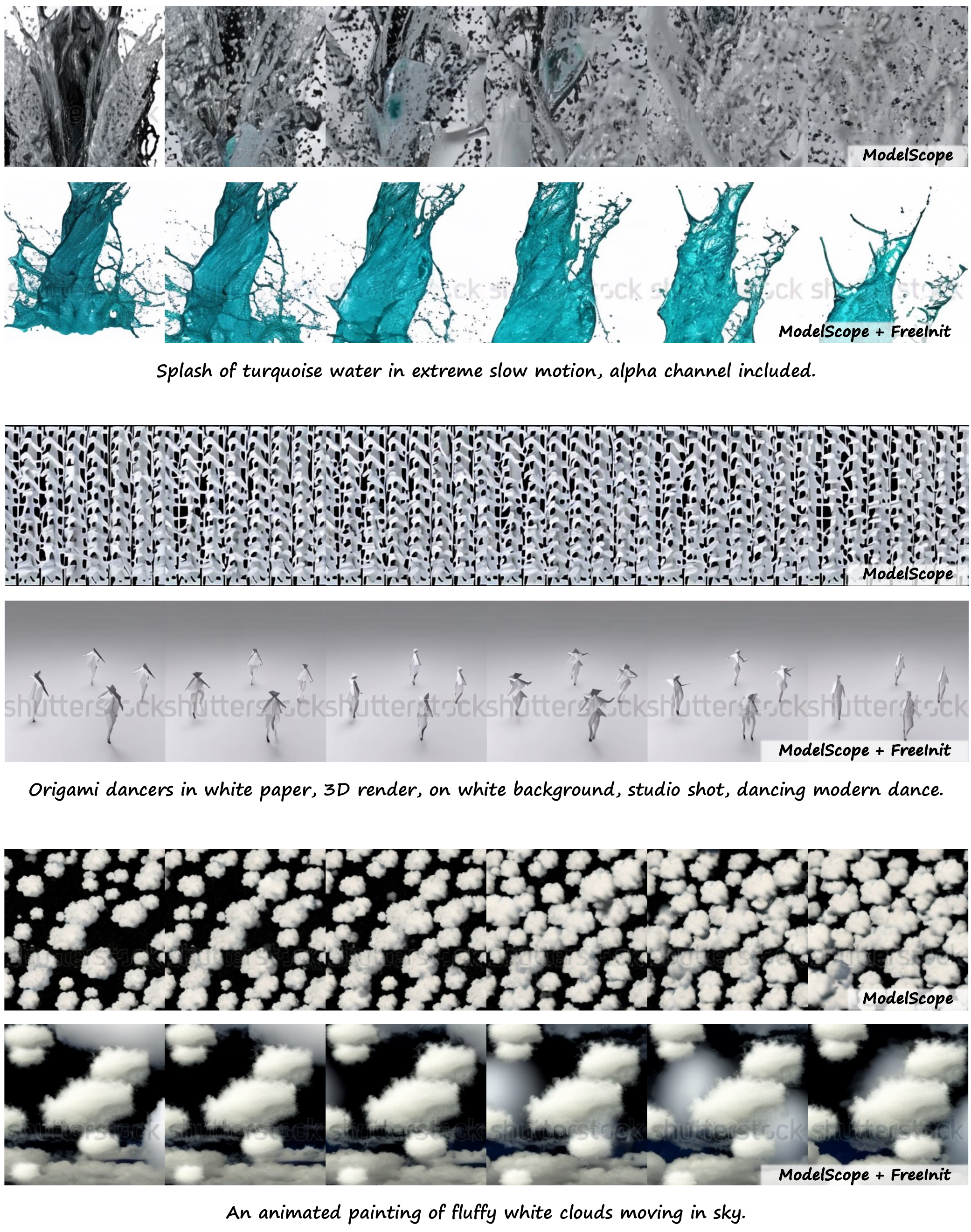}
   \end{center}
   \caption{\textbf{More Qualitative Results of FreeInit on ModelScope.}}
   \label{fig:modelscope_1}
\end{figure*}

\begin{figure*}[t]
   \begin{center}
      \includegraphics[width=0.98\linewidth]{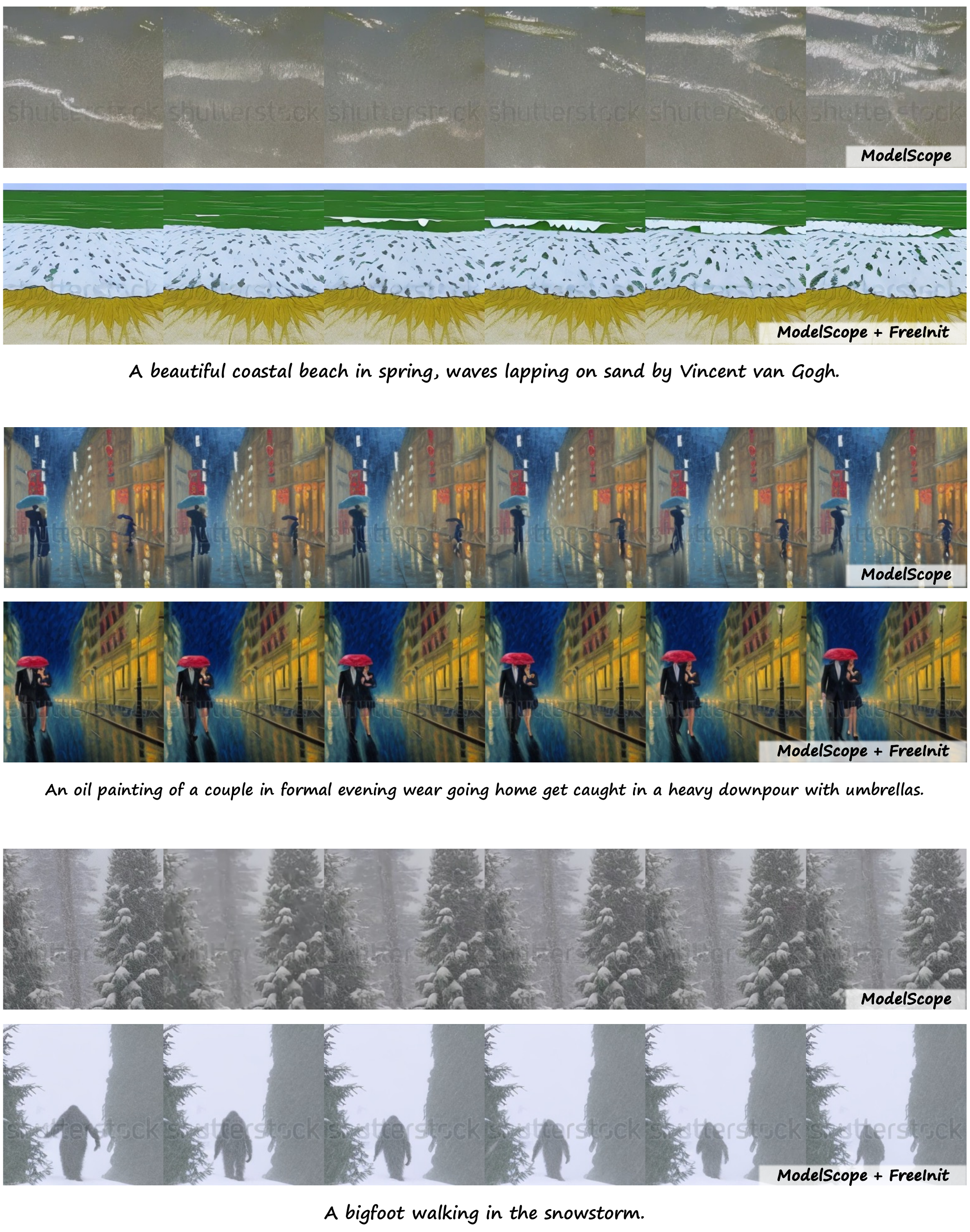}
   \end{center}
   \caption{\textbf{More Qualitative Results of FreeInit on ModelScope.}}
   \label{fig:modelscope_2}
\end{figure*}

\begin{figure*}[t]
   \begin{center}
      \includegraphics[width=0.98\linewidth]{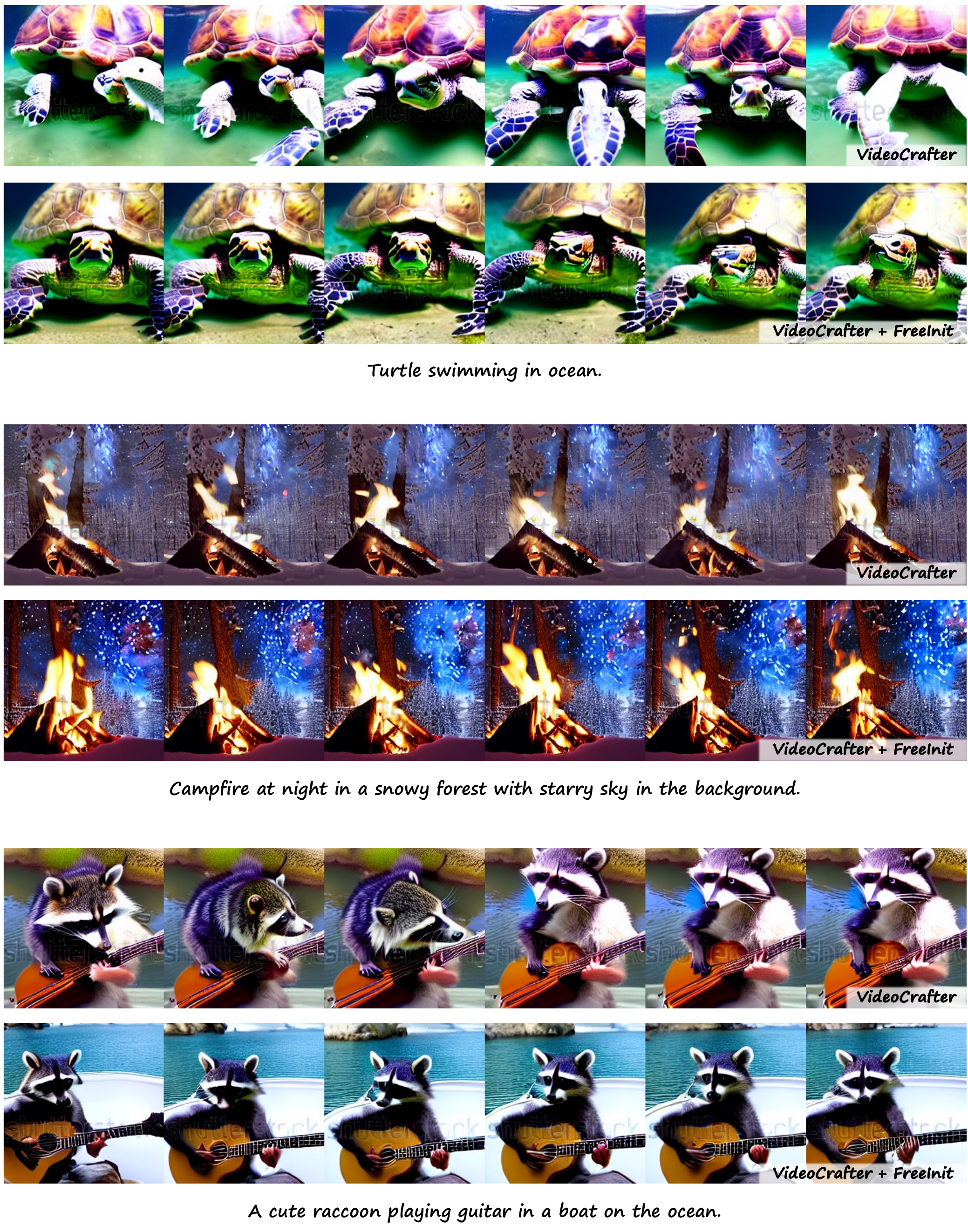}
   \end{center}
   \caption{\textbf{More Qualitative Results of FreeInit on VideoCrafter.}}
   \label{fig:videocrafter_1}
\end{figure*}

\begin{figure*}[t]
   \begin{center}
      \includegraphics[width=0.98\linewidth]{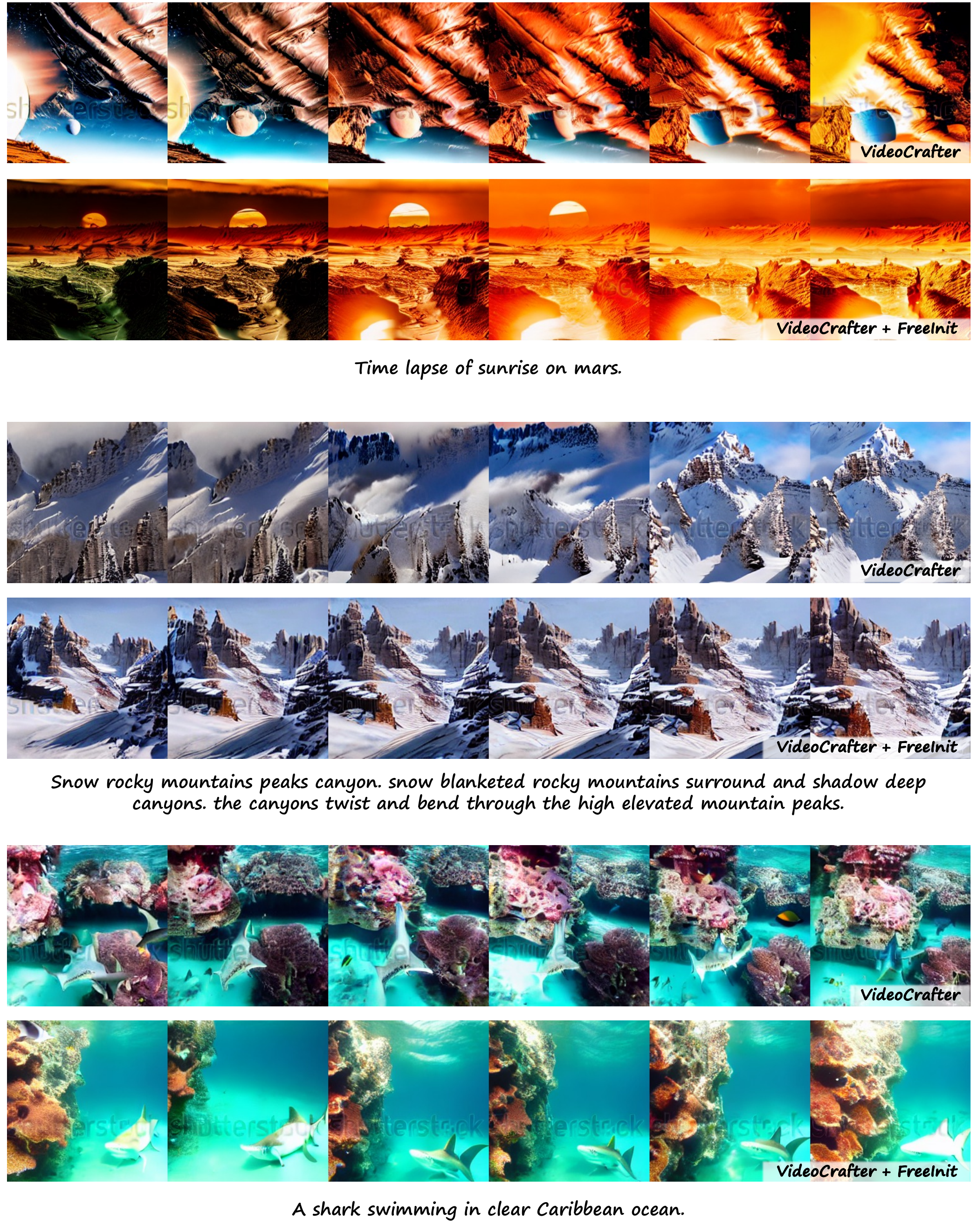}
   \end{center}
   \caption{\textbf{More Qualitative Results of FreeInit on VideoCrafter.}}
   \label{fig:videocrafter_2}
\end{figure*}

\begin{figure*}[t]
   \begin{center}
      \includegraphics[width=0.98\linewidth]{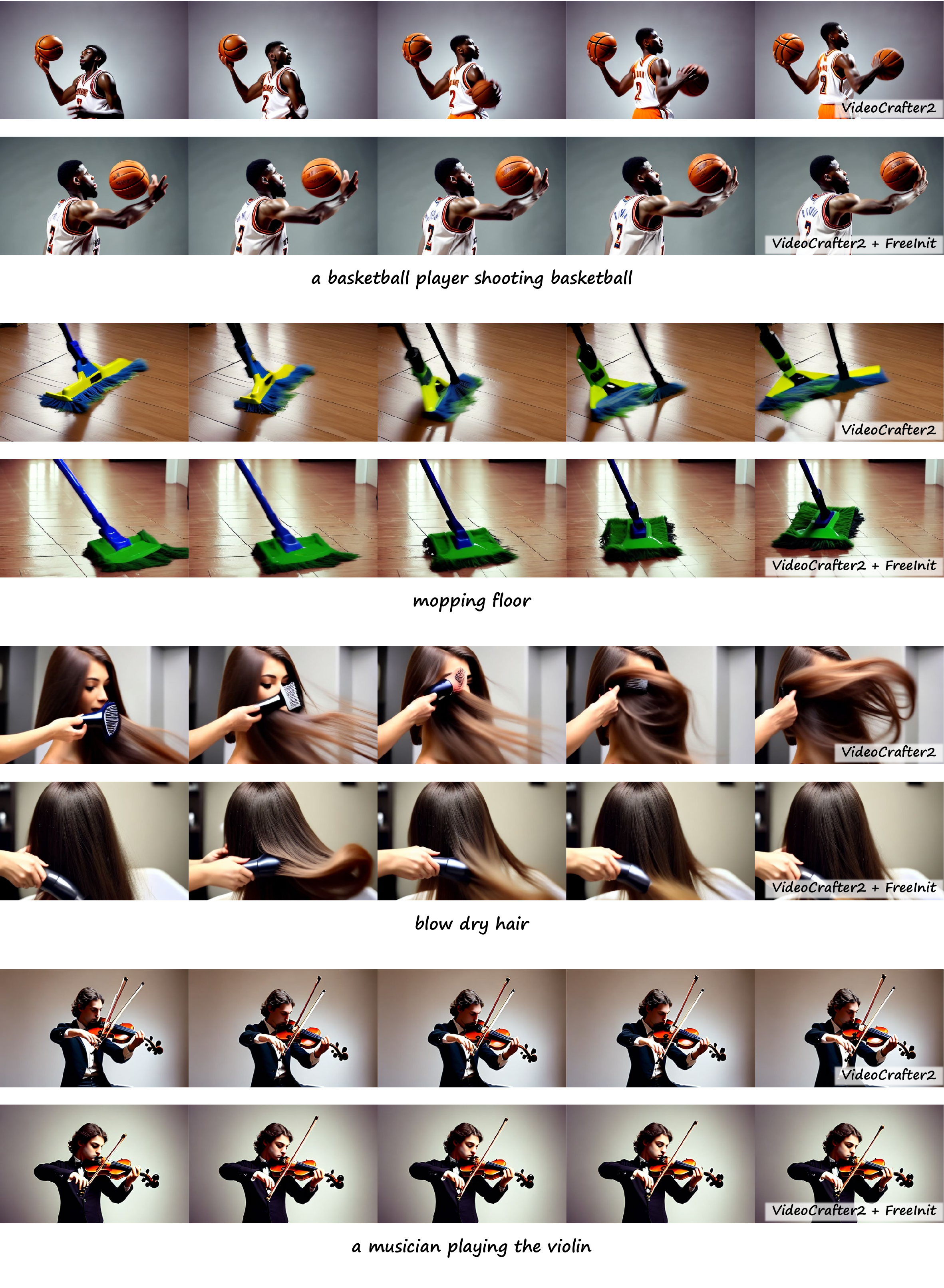}
   \end{center}
   \caption{\textbf{More Qualitative Results of FreeInit on VideoCrafter2.}}
   \label{fig:videocrafter2}
\end{figure*}

\noindent\textbf{Failure Cases.} In some cases where the video includes small and fast-moving foreground objects, performing FreeInit occasionally leads to the distinction of the object. As shown in the example \cref{fig:failure}, although the temporal consistency of the rose is enhanced, the small object ``waterdrop'' almost vanishes. This is because the iterative low-frequency refinement strategy tends to guide the generation towards the more stable low-frequency subjects. As the iteration progresses, the enhanced semantics within the initial noise's lower frequencies increasingly take more control of the generation process, causing a partial loss of control from the text condition. Regarding this, the users can choose to use fewer FreeInit iterations, tune the frequency filter parameters, or adjust the classifier-free guidance weight to balance the trade-off between depicting small objects and enhancing temporal consistency, according to their specific needs and preferences.

\section{Potential Negative Societal Impacts}
\label{sec:social}
FreeInit is a research focused on improving the inference quality of existing video diffusion models without favoring specific content categories. Nonetheless, its application in aiding other video generation models could potentially be exploited for malicious purposes, resulting in the creation of fake content.

\end{document}